\documentclass[letterpaper]{article}
\usepackage{aaai2027}
\usepackage[hyphens]{url}
\usepackage{graphicx}
\usepackage{natbib}
\usepackage{caption}
\usepackage{amsmath,amssymb}
\usepackage{booktabs}
\usepackage{multirow}
\usepackage{makecell}
\usepackage[table]{xcolor}
\usepackage{colortbl}
\usepackage{algorithm}
\usepackage{algorithmic}
\usepackage{enumitem}
\usepackage{placeins}
\usepackage[capitalize,noabbrev]{cleveref}

\pdfmapfile{+newtx.map}
\pdfmapfile{+utm.map}
\pdfmapfile{+uhv.map}
\pdfmapfile{+ucr.map}
\pdfmapfile{+symbols.map}
\pdfmapfile{+cm.map}
\pdfmapfile{+cmextra.map}
\pdfmapfile{+local8r.map}
\pdfmapfile{+qtm.map}
\pdfmapfile{+qtm-ts1.map}

\definecolor{cmvfrow}{RGB}{215,235,210}      
\definecolor{winrow}{RGB}{219,237,214}   
\definecolor{secrow}{RGB}{238,238,238}   
\definecolor{gainred}{RGB}{200,30,30}        
\definecolor{stdgray}{RGB}{120,120,120}      
\newcommand{\stdsub}[1]{\textcolor{stdgray}{\scriptsize$_{\pm#1}$}}
\newcommand{\dgain}[1]{{\,\textcolor{gainred}{\scriptsize$\uparrow$#1}}}

\definecolor{acceptgreen}{RGB}{34,139,34}
\definecolor{weakgray}{RGB}{120,120,120}
\newcommand{\addt}[1]{\textcolor{acceptgreen}{#1}}              
\newcommand{\weakp}[1]{\textcolor{weakgray}{\textit{#1}}}        
\definecolor{mProTeGi}{RGB}{93,171,102}      
\definecolor{mOPRO}{RGB}{139,77,168}         
\definecolor{mDSPy}{RGB}{122,92,68}          
\definecolor{mTG}{RGB}{79,169,217}           
\definecolor{mRE}{RGB}{242,162,58}           
\definecolor{mCMVF}{RGB}{224,74,110}         
\newcommand{\meth}[2]{\textbf{\textcolor{#1}{#2}}}                
\newcommand{\methcmvf}{\meth{mCMVF}{CMVF}}
\definecolor{rowmed}{RGB}{250,235,235}      
\definecolor{rowchart}{RGB}{233,240,250}    
\definecolor{rowreal}{RGB}{233,246,235}     
\definecolor{badgemed}{RGB}{195,70,70}
\definecolor{badgechart}{RGB}{55,105,200}
\definecolor{badgereal}{RGB}{45,150,75}

\newcommand{\modelIcon}[1]{%
  \raisebox{-0.32\height}{\includegraphics[height=11pt,keepaspectratio]{figures/#1}}%
}
\newcommand{\iconQwenA}{\modelIcon{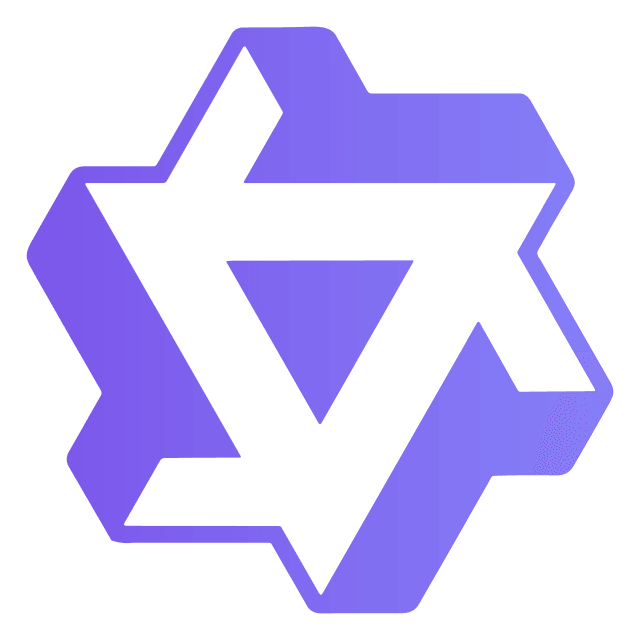}}      
\newcommand{\iconQwenB}{\modelIcon{qwen-color.png}}      
\newcommand{\iconLLaVA}{\modelIcon{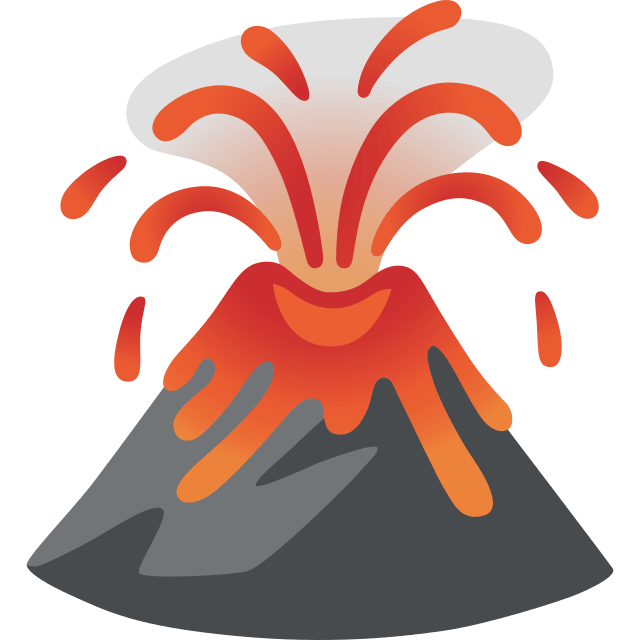}}     
\newcommand{\iconPhi}{\modelIcon{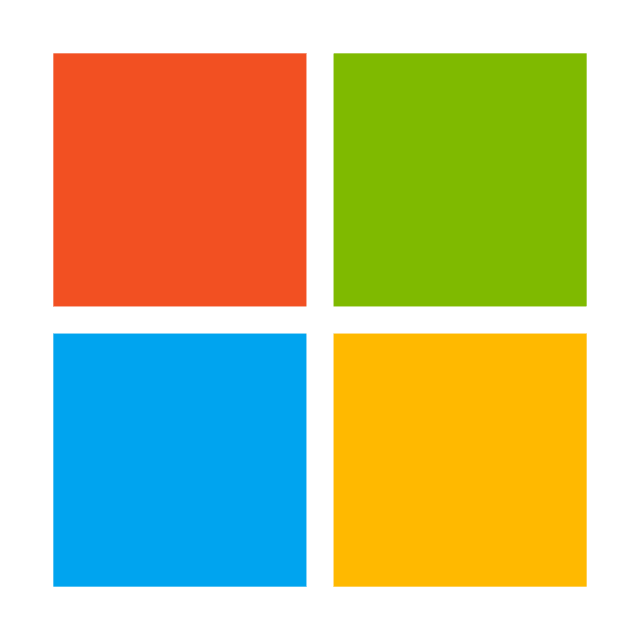}}   
\newcommand{\modelTag}[2]{#1\,\textbf{#2}}

\setlist[itemize]{leftmargin=1.4em,topsep=2pt,itemsep=2pt,parsep=0pt}
\setlist[enumerate]{leftmargin=1.6em,topsep=2pt,itemsep=2pt,parsep=0pt}

\title{Are Prompt Optimizers Blind? Cross-Modal Visual Feedback for Automatic Prompt Optimization}

\author{
    Haoyue Liu\textsuperscript{\rm 1}, Xiaoyu Ma\textsuperscript{\rm 1}, Ye Chen\textsuperscript{\rm 2}, Yuexian Zou\textsuperscript{\rm 3,4}, Xiaoying Tang\textsuperscript{\rm 1}\corresponding
}
\affiliations{
    \textsuperscript{\rm 1} The Chinese University of Hong Kong, Shenzhen, China\\
    \textsuperscript{\rm 2} Xi'an Jiaotong University, China\\
    \textsuperscript{\rm 3} Peking University, China\\
    \textsuperscript{\rm 4} Guangdong Provincial Key Laboratory of Ultra High Definition Immersive Media Technology, China\\
}

\begin{document}

\nocopyright

\maketitle

\begin{abstract}
\sloppy
Automatic prompt optimization (APO) has been widely adopted to adapt vision-language models (VLMs) to downstream tasks without weight updates, yielding promising results. However, on multimodal tasks the effectiveness of APO is fundamentally bottlenecked by a \emph{blind feedback channel}: the optimizer reads the question, the prediction, and the gold answer, but never the input image on which the model failed, and therefore cannot diagnose visually grounded errors. As a remedy, we introduce \textbf{Cross-Modal Visual Feedback (CMVF)}. CMVF incorporates (1) a failure-conditioned visual diagnosis stage, in which a stronger optimizer VLM inspects each failed image \emph{without access to predictions or labels}, and (2) an error-aware aggregation stage that compresses these observations into reusable, task-level visual blind-spot patterns that drive the prompt rewrite. Crucially, the image is consumed \emph{only} during optimization; the deployed artifact is an ordinary text prompt that runs at the same inference cost as any text-only baseline. Extensive results across 12 VQA datasets and 4 target VLMs demonstrate that CMVF consistently ranks first, improving over the strongest baseline on every target by \textbf{+2.4} points on average (up to \textbf{+6.5} on individual benchmarks), while the optimizer self-organizes into expert-style visual checklists that transfer across models without re-optimization. Our code and optimized prompts will be released.

\end{abstract}

\section{Introduction}
\label{sec:intro}

Vision-language models (VLMs) are the default substrate for visual question answering, medical imaging, chart understanding, and document interpretation~\cite{masry2022chartqa,mathew2021docvqa}, yet their accuracy is notoriously sensitive to the wording of the textual prompt---on our benchmarks, mean accuracy on Qwen3.5-4B varies from 70.3\% to 75.0\% across competitive prompts alone. Automatic prompt optimization (APO)---spanning \emph{evolutionary search}~\cite{fernando2023promptbreeder,guo2024connecting}, \emph{textual-gradient methods}~\cite{pryzant2023automatic,yuksekgonul2025optimizing}, \emph{response-trajectory tracking}~\cite{zhang2024revolve}, and \emph{declarative program compilation}~\cite{khattab2024dspy}---has become an effective, weight-free way to adapt these models, yielding promising results on text-centric tasks.

\begin{figure}[t]
\centering
\includegraphics[width=\columnwidth]{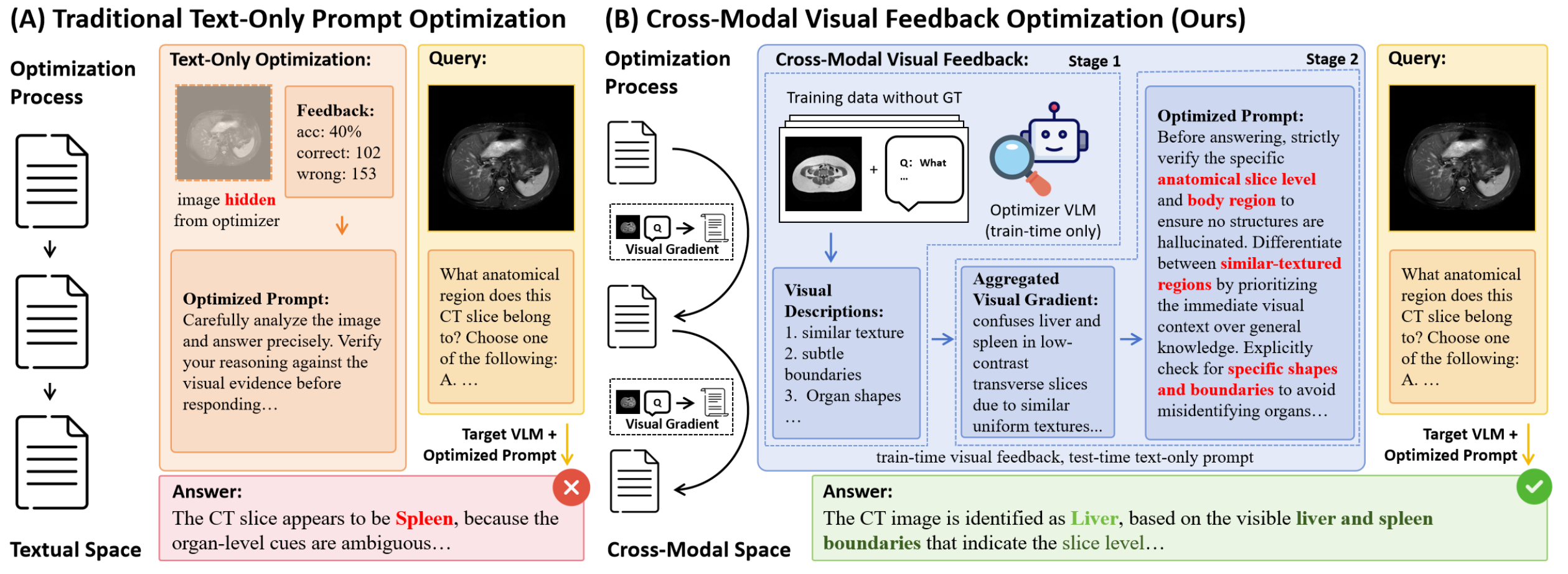}
\caption{Motivation for CMVF. Existing text-only APO sees scores and error triples
but misses the visual causes of failures. CMVF uses failed images only during
optimization to aggregate visual blind spots into a reusable text prompt, with no
extra visual-feedback calls at test time.}
\label{fig:motivation}
\end{figure}

However, when applied to multimodal tasks these optimizers all share the same structural gap: their feedback channel is \emph{blind}. The optimizer sees the question, the prediction, and the gold answer, but never the image on which the target failed~(\Cref{fig:motivation})---like a teacher grading an exam without ever seeing the figures. It can register \emph{that} an error occurred but not \emph{why}: a confused anatomical slice level, an overlooked axis label, or a global scene prior that overrides local evidence. We argue that the bottleneck of multimodal APO is not the optimization machinery but the \emph{evidence} available to it.

As a remedy, we introduce \textbf{Cross-Modal Visual Feedback (CMVF)}, built on a simple principle: \emph{optimization feedback must have access to the modality in which failures originate.} CMVF lets a stronger optimizer VLM inspect the failed images during optimization---first describing question-relevant visual evidence without seeing the prediction or ground-truth answer, then aggregating these descriptions with the error triples into reusable visual blind spots that drive the prompt rewrite. Unlike multimodal prompt optimization (MPO)~\cite{choi2025multimodal}, which jointly optimizes a $(\text{text},\text{image})$ prompt pair and keeps the image in the loop at deployment, CMVF pays for the visual signal \emph{only once}, at optimization time, and compiles it into a purely textual prompt; deployment therefore adds no VLM calls and reuses the same serving interface as any text-only prompt.

Our contributions are three-fold:
\begin{enumerate}[label=\textbf{\arabic*.},leftmargin=*]
    \item \textbf{Problem.} We identify and characterize the \emph{visual-feedback blindness} shared by all text-only APO methods on multimodal tasks, and show the bottleneck is \emph{evidence}, not optimization machinery.
    \item \textbf{Method.} We propose CMVF, which incorporates (1)~failure-conditioned, label-free visual diagnosis and (2)~error-aware aggregation into reusable visual blind-spot patterns, realizing a \emph{train-time-multimodal, deploy-time-text-only} design at zero added inference cost.
    \item \textbf{Results \& analysis.} Across 12 datasets and 4 target VLMs, CMVF ranks first everywhere, beating the strongest baseline by $+2.4$\,pp on average (up to $+6.5$\,pp); token-budget-matched ablations attribute the gain to visual perception, the optimizer self-organizes into \emph{expert-style visual checklists}, and these strategies transfer across models without re-optimization.
\end{enumerate}

\section{Cross-Modal Visual Feedback}
\label{sec:method}

The preceding discussion raises two questions that define the design of CMVF. \emph{First}, how can a prompt optimizer obtain visual diagnostic information without leaking the ground-truth answer into the visual analysis? \emph{Second}, how can this visual information improve the deployed prompt without adding any inference-time VLM calls? CMVF answers these two questions with a two-stage visual-gradient channel (\Cref{sec:visual_grad}) and a fusion-update loop that preserves a standard textual-prompt inference interface (\Cref{sec:fusion}).

\subsection{Problem Setup}
\label{sec:setup}

Given a target VLM $\mathcal{M}_T$ and training data $\mathcal{D}=\{(x_i,q_i,y_i)\}_{i=1}^N$ ($x_i$ image, $q_i$ question, $y_i$ ground-truth answer), we seek a text prompt $p^*\in\mathcal{P}$ that maximizes task accuracy:
\begin{equation}
    p^*=\arg\max_{p\in\mathcal{P}}\frac{1}{N}\sum_{i=1}^{N}\mathbf{1}[\mathcal{M}_T(p,x_i,q_i)=y_i].
\end{equation}
Following TextGrad~\cite{yuksekgonul2025optimizing}, we optimize two prompts jointly: $p_{yn}$ for yes/no questions and $p_{oe}$ for open-ended questions, both with a \texttt{\{fs\}} placeholder for few-shot examples. At step $t\!\in\!\{1,\dots,T\}$, let $\hat{y}_i\!:=\!\mathcal{M}_T(p_t,x_i,q_i)$ be the prediction under the current prompt $p_t=(p_{yn}^{(t)},p_{oe}^{(t)})$; we form the correct set $\mathcal{C}_t$ and the wrong set $\mathcal{W}_t=\{(x_i,q_i,\hat{y}_i,y_i):\hat{y}_i\neq y_i\}$. A validation set $\mathcal{V}$ selects the best prompt across steps, and if a rewrite drops the \texttt{\{fs\}} placeholder we restore the validation-best prompt.

\subsection{The Visual-Feedback Bottleneck}
\label{sec:textgrad}

Standard APO constructs feedback from accuracy and textual error triples, $g_t^{\text{text}}=\big(\text{acc}(p_t),\mathcal{E}_t^{\text{sub}}\big)$ with $\mathcal{E}_t^{\text{sub}}\!\subseteq\!\{(q_i,\hat{y}_i,y_i)\}$. This exposes \emph{which} questions failed but omits the visual evidence that caused them: a medical VQA model may confuse adjacent anatomical slice levels, a chart model may miss a small axis label, a real-world model may rely on a global scene prior. The textual triple sees the symptom; the cause stays hidden. CMVF closes this loop with the two-stage visual channel below.

\subsection{Visual Diagnostic Channel}
\label{sec:visual_grad}

\begin{figure*}[t]
\centering
\includegraphics[width=0.85\textwidth]{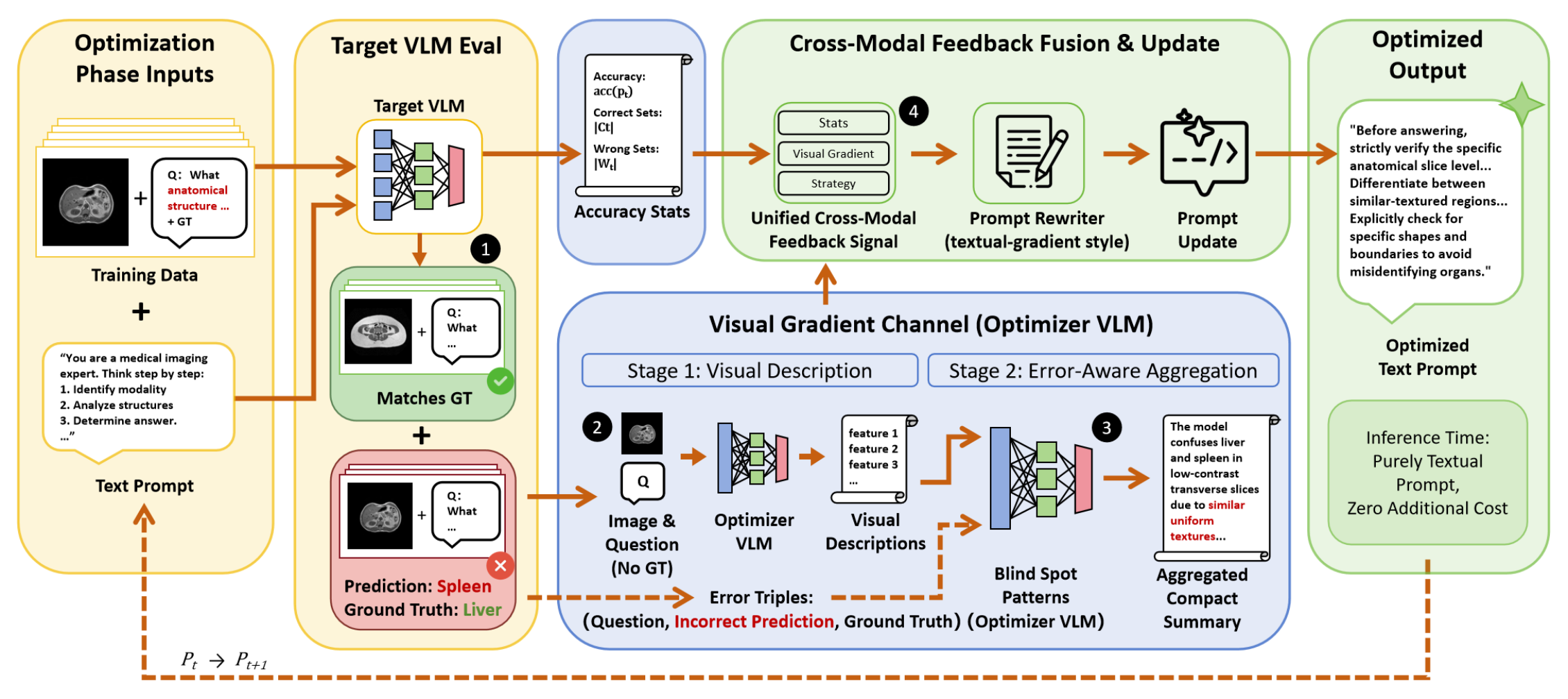}
\caption{CMVF optimization pipeline. The target VLM is evaluated on training
examples; the optimizer VLM inspects failed images, aggregates visual blind spots,
and rewrites the text prompt. Inference uses the optimized text prompt with no
extra visual-feedback calls.}
\label{fig:framework}
\end{figure*}

CMVF adds a visual-gradient channel to the feedback loop. During optimization only, a stronger optimizer VLM $\mathcal{M}_O$ inspects failed images and summarizes recurring visual blind spots. \Cref{fig:framework} gives the overview.

\paragraph{Stage 1: question-aware visual description.}
For every wrong example, $\mathcal{M}_O$ receives only the image and question and produces a question-aware visual description:
\begin{equation}
    v_i=\mathcal{M}_O(x_i,q_i;\pi_{\text{vis}}), \quad \forall (x_i,q_i,\hat{y}_i,y_i)\in\mathcal{W}_t,
\end{equation}
where $\pi_{\text{vis}}$ is a fixed Stage-1 meta-prompt asking for question-relevant visual evidence (supplementary material). Crucially, the prediction $\hat{y}_i$ and ground truth $y_i$ are \emph{withheld} at this stage. This prevents answer-conditioned rationalization---the optimizer cannot manufacture a post-hoc explanation for a label it has not yet seen---and forces it to record perception-level evidence rather than rationalize a known label.

\paragraph{Stage 2: error-aware aggregation.}
Individual descriptions are noisy and idiosyncratic. We therefore pair each $v_i$ with its error triple and ask $\mathcal{M}_O$ to summarize recurring blind spots:
\begin{equation}
    g_t^{\text{vis}}=\text{Agg}\big(\{(q_i,\hat{y}_i,y_i,v_i):(x_i,q_i,\hat{y}_i,y_i)\in\mathcal{W}_t\}\big).
\end{equation}
The aggregator receives at most 30 wrong cases per step with field-level truncation. The two-stage decomposition separates visual perception from supervised error attribution: Stage~1 records \emph{what is visible}; Stage~2 identifies \emph{which visual patterns repeatedly correlate with mistakes}, compressing them into a compact set of reusable diagnostic instructions (e.g., ``check small text,'' ``verify anatomical slice level,'' ``avoid global scene priors''). This compression is what makes broad inspection at Stage~1 affordable: many descriptions in, one prompt-sized signal out.

\subsection{Fusion, Prompt Update, and Inference}
\label{sec:fusion}

\paragraph{Fused feedback.}
We fuse accuracy statistics and visual blind spots into a unified feedback string:
\begin{equation}
\label{eq:fusion}
    g_t^{\text{CM}}=\textsc{Concat}\big(\text{acc}(p_t),|\mathcal{C}_t|,|\mathcal{W}_t|,g_t^{\text{vis}},s\big),
\end{equation}
where $s$ instructs the optimizer to rewrite the prompt so the target model explicitly checks the identified visual features before answering. Because $g_t^{\text{vis}}$ is constructed from both visual descriptions and textual error triples, CMVF \emph{augments} rather than replaces the existing textual error signal.

\paragraph{Update and inference.}
The prompt is updated by $p_{t+1}=\textsc{Update}(p_t,g_t^{\text{CM}})$. The default \textsc{Update} is a TextGrad-style rewrite~\cite{yuksekgonul2025optimizing}; CMVF$^*$ swaps in a REVOLVE-style alternative~\cite{zhang2024revolve} on the same visual-gradient pipeline, isolating whether gains come from the visual channel rather than a particular rewriter. At deployment, the target VLM receives only the image, question, and optimized text prompt $(p_{yn},p_{oe})$---no additional visual analysis at inference (\Cref{alg:cmvf}).

\begin{algorithm}[t]
\caption{CMVF Prompt Optimization}
\label{alg:cmvf}
\begin{algorithmic}[1]
\REQUIRE Target VLM $\mathcal{M}_T$, optimizer VLM $\mathcal{M}_O$, training set $\mathcal{D}$, validation set $\mathcal{V}$, steps $T$
\STATE Initialize prompts $(p_{yn},p_{oe})$ and validation-best prompts
\FOR{$t=1$ to $T$}
    \STATE Evaluate $\mathcal{M}_T$ on $\mathcal{D}$ to obtain $\mathcal{C}_t$ and $\mathcal{W}_t$
    \STATE Evaluate on $\mathcal{V}$ and update the validation-best prompts if improved
    \FOR{each $(x_i,q_i,\hat{y}_i,y_i)\in\mathcal{W}_t$}
        \STATE $v_i \leftarrow \mathcal{M}_O(x_i,q_i;\pi_{\text{vis}})$ \COMMENT{image + question only}
    \ENDFOR
    \STATE $g_t^{\text{vis}}\leftarrow \textsc{Agg}(\{(q_i,\hat{y}_i,y_i,v_i):(x_i,q_i,\hat{y}_i,y_i)\in\mathcal{W}_t\})$
    \STATE $g_t^{\text{CM}}\leftarrow \textsc{Concat}(\text{acc}(p_t),|\mathcal{C}_t|,|\mathcal{W}_t|,g_t^{\text{vis}},s)$
    \STATE $(p_{yn},p_{oe})\leftarrow \textsc{Update}((p_{yn},p_{oe}),g_t^{\text{CM}})$
    \STATE Restore validation-best prompts if \texttt{\{fs\}} is lost
\ENDFOR
\RETURN validation-best prompts
\end{algorithmic}
\end{algorithm}

\subsection{Modeling Intuition}
\label{sec:theory}

The role of visual feedback can be understood through a simple error-cause decomposition. Let $\alpha_V\in[0,1]$ denote the fraction of currently wrong examples whose failure has a visual component, and let $I_V\in[0,1]$ denote how informative the optimizer's visual analysis is about these causes. Feedback without visual inspection carries only indirect visual hints; denote its residual visual signal by $\delta_T$. With error rate $L(p)\!:=\!1{-}\text{acc}(p)$ and per-step reduction $\Delta L\!:=\!L(p_t){-}L(p_{t+1})$, under an additive view of textual and visual error sources the expected advantage of cross-modal feedback scales as
\begin{equation}
\mathbb{E}\big[\Delta L_{\text{CM}}-\Delta L_{\text{text}}\big]\;\gtrsim\;\alpha_V\,I_V\;-\;\delta_T.
\label{eq:intuition}
\end{equation}
This intuition (not a convergence guarantee) predicts CMVF should help most when $\alpha_V$ is large (medical VQA, TextVQA OCR, RealWorldQA) and least when $\alpha_V\!\to\!0$; both match our empirical pattern in \Cref{sec:main_results}. This also clarifies the role of Stage-2 aggregation. A single visual description may be noisy or instance-specific, so its information about the underlying visual cause is limited. By aggregating many failed cases, CMVF suppresses instance-specific noise, surfaces recurring visual patterns, and converts per-image observations into a compact prompt-level signal. In this sense, aggregation increases the \emph{effective} $I_V$, which is consistent with the ablation results in \Cref{sec:ablation}.

\FloatBarrier
\section{Experiments}
\label{sec:experiments}

In this section, we run experiments to address five questions:
\begin{itemize}[nosep,leftmargin=*]
    \item \textbf{Q1}: Does CMVF beat APO baselines that do not incorporate visual feedback, across multiple target VLMs? (\Cref{sec:main_results})
    \item \textbf{Q2}: Does the optimizer learn anything \emph{interpretable} about the failure modes? (\Cref{sec:emergent})
    \item \textbf{Q3}: Do the optimized strategies \emph{transfer} across target models without re-optimization? (\Cref{sec:transfer})
    \item \textbf{Q4}: Where do the gains come from---image-aware perception, or extra feedback tokens? (\Cref{sec:ablation})
    \item \textbf{Q5}: Does CMVF generalize and pay an acceptable training cost with unchanged inference? (\Cref{sec:efficiency_generalization})
\end{itemize}
Stage-2 variants (CM-Mem, Diff), dataset details, prompt templates, and case studies are in the Appendix.

\subsection{Setup}
\label{sec:exp_setup}

\paragraph{Models and benchmarks.}
We use Qwen3.5-9B as the optimizer VLM $\mathcal{M}_O$ and evaluate four target VLMs: Qwen3.5-4B, Qwen2.5-VL-7B~\cite{wang2024qwen2vl}, LLaVA-1.6-Mistral-7B~\cite{li2023llava}, and Phi-3.5-vision~\cite{abdin2024phi3}. The suite covers 12 datasets: medical VQA datasets SLAKE CT/MRI/X-Ray, VQA-RAD, and PathVQA~\cite{liu2021slake,lau2018dataset,he2020pathvqa}; reasoning and chart/OCR datasets ScienceQA, ChartQA, TextVQA, and MMMU~\cite{lu2022learn,masry2022chartqa,singh2019towards,yue2024mmmu}; and RealWorldQA, AI2D, and DocVQA~\cite{xai2024realworldqa,kembhavi2016diagram,mathew2021docvqa}. We focus the main comparison on eight benchmarks and report the remaining four (ScienceQA, AI2D, DocVQA, MMMU) separately in \Cref{tab:official_rescored}.

\paragraph{Baselines and protocol.}
We compare against ProTeGi~\cite{pryzant2023automatic}, OPRO~\cite{yang2024large}, DSPy~\cite{khattab2024dspy}, TextGrad (TG)~\cite{yuksekgonul2025optimizing}, and REVOLVE (RE)~\cite{zhang2024revolve}. Each benchmark uses 150 training samples, 150 validation samples, and the full test set. All methods share data splits, initialization prompts, the same optimizer VLM, and 20 optimization steps. Results are averaged over 3 runs. Evaluation follows each benchmark's official scoring protocol (exact match after lowercasing and punctuation removal, plus a small set of domain synonyms, e.g., ``x-ray''/``xray'').

\subsection{A1: CMVF Ranks First on All Four Target VLMs}
\label{sec:main_results}

\begin{table*}[t]
\centering
\scriptsize
\setlength{\tabcolsep}{2.6pt}
\renewcommand{\arraystretch}{0.88}
\begin{tabular*}{\textwidth}{@{\extracolsep{\fill}} l l | c c c c c | c c c | c}
\toprule
\multirow{2}{*}{\textbf{Model}} & \multirow{2}{*}{\textbf{Method}}
& \multicolumn{5}{c|}{\textbf{Medical VQA}\,$\uparrow$}
& \multicolumn{3}{c|}{\textbf{General VQA}\,$\uparrow$}
& \multirow{2}{*}{\textbf{Mean}\,$\uparrow$} \\
\cmidrule(lr){3-7} \cmidrule(lr){8-10}
& & \textbf{CT} & \textbf{MRI} & \textbf{X-Ray} & \textbf{RAD} & \textbf{Path}
& \textbf{Chart} & \textbf{Text} & \textbf{RWQA} & \\
\midrule

\multirow{7}{*}{\modelTag{\iconQwenA}{Qwen3.5-4B}}
& ProTeGi  & 75.3\stdsub{2.8} & 78.1\stdsub{0.9} & 81.7\stdsub{1.4} & 59.4\stdsub{3.1} & 45.4\stdsub{0.5} & 78.0\stdsub{3.4} & 65.2\stdsub{2.3} & 80.4\stdsub{1.9} & 70.4 \\
& OPRO     & 73.2\stdsub{3.4} & 67.7\stdsub{6.1} & 81.8\stdsub{1.0} & 55.2\stdsub{4.7} & 44.6\stdsub{1.9} & 25.1\stdsub{28.4} & 69.1\stdsub{0.6} & 81.7\stdsub{3.3} & 62.3 \\
& \cellcolor{secrow}DSPy     & \cellcolor{secrow}\underline{77.6}\stdsub{1.1} & \cellcolor{secrow}77.7\stdsub{1.4} & \cellcolor{secrow}80.7\stdsub{1.3} & \cellcolor{secrow}\underline{62.1}\stdsub{0.8} & \cellcolor{secrow}\underline{47.5}\stdsub{2.8} & \cellcolor{secrow}\textbf{81.6}\stdsub{1.9} & \cellcolor{secrow}66.3\stdsub{2.2} & \cellcolor{secrow}82.6\stdsub{0.9} & \cellcolor{secrow}\underline{72.0} \\
& TG       & 75.9\stdsub{0.3} & 75.3\stdsub{1.3} & 80.7\stdsub{1.6} & 57.9\stdsub{3.1} & 45.4\stdsub{2.0} & 78.6\stdsub{2.7} & 63.8\stdsub{1.4} & \underline{84.8}\stdsub{0.7} & 70.3 \\
& RE       & 75.5\stdsub{0.6} & \underline{79.0}\stdsub{0.1} & \underline{83.8}\stdsub{0.5} & 59.2\stdsub{2.0} & 44.3\stdsub{1.1} & 75.1\stdsub{2.7} & 64.3\stdsub{1.6} & 80.9\stdsub{2.2} & 70.3 \\
& CMVF$^{\!*}$ & 76.8\stdsub{0.4} & 77.9\stdsub{2.2} & 84.1\stdsub{0.2} & 61.3\stdsub{1.4} & 46.9\stdsub{0.8} & 78.8\stdsub{2.0} & 65.7\stdsub{0.8} & 83.3\stdsub{1.8} & 71.9 \\
& \cellcolor{winrow}\textbf{CMVF} & \cellcolor{winrow}\textbf{77.9}\stdsub{1.3} & \cellcolor{winrow}\textbf{80.7}\stdsub{3.1} & \cellcolor{winrow}\textbf{84.5}\stdsub{1.2} & \cellcolor{winrow}\textbf{63.4}\stdsub{1.0} & \cellcolor{winrow}\textbf{49.0}\stdsub{2.8} & \cellcolor{winrow}\underline{81.4}\stdsub{2.0} & \cellcolor{winrow}\textbf{71.5}\stdsub{2.7} & \cellcolor{winrow}\textbf{91.3}\stdsub{3.8} & \cellcolor{winrow}\textbf{75.0}\dgain{3.0} \\
\midrule

\multirow{7}{*}{\modelTag{\iconQwenB}{Qwen2.5-VL-7B}}
& ProTeGi & 64.0\stdsub{1.3} & 59.4\stdsub{4.9} & \underline{73.5}\stdsub{1.1} & 47.5\stdsub{2.3} & 37.9\stdsub{0.9} & 56.3\stdsub{0.6} & 59.8\stdsub{0.6} & 58.3\stdsub{11.0} & 57.1 \\
& OPRO & 60.1\stdsub{6.4} & 61.4\stdsub{8.3} & 69.4\stdsub{7.5} & 42.6\stdsub{0.1} & 29.5\stdsub{4.4} & 54.2\stdsub{3.8} & \underline{60.0}\stdsub{0.2} & 50.4\stdsub{8.0} & 53.4 \\
& DSPy & 60.8\stdsub{1.1} & 61.0\stdsub{2.8} & 72.1\stdsub{1.3} & 47.0\stdsub{0.6} & 32.7\stdsub{1.3} & \textbf{59.4}\stdsub{2.1} & 55.5\stdsub{2.6} & 60.0\stdsub{8.6} & 56.1 \\
& \cellcolor{secrow}TG & \cellcolor{secrow}\underline{64.5}\stdsub{2.0} & \cellcolor{secrow}\underline{69.2}\stdsub{1.3} & \cellcolor{secrow}\underline{73.5}\stdsub{1.6} & \cellcolor{secrow}48.8\stdsub{1.6} & \cellcolor{secrow}36.8\stdsub{1.9} & \cellcolor{secrow}54.4\stdsub{1.9} & \cellcolor{secrow}59.4\stdsub{1.8} & \cellcolor{secrow}\underline{61.9}\stdsub{2.7} & \cellcolor{secrow}\underline{58.6} \\
& RE & 63.5\stdsub{0.2} & \textbf{69.4}\stdsub{0.6} & 72.1\stdsub{0.2} & 48.3\stdsub{1.0} & \underline{38.6}\stdsub{0.9} & \underline{57.5}\stdsub{0.3} & 57.1\stdsub{3.7} & 58.7\stdsub{2.1} & 58.2 \\
& CMVF$^{\!*}$ & 63.8\stdsub{0.9} & 68.7\stdsub{3.2} & 73.2\stdsub{2.2} & \underline{49.9}\stdsub{1.4} & 35.4\stdsub{2.7} & 55.7\stdsub{1.2} & 59.3\stdsub{1.5} & \underline{61.9}\stdsub{0.7} & 58.5 \\
& \cellcolor{winrow}\textbf{CMVF} & \cellcolor{winrow}\textbf{65.0}\stdsub{1.1} & \cellcolor{winrow}\textbf{69.4}\stdsub{2.3} & \cellcolor{winrow}\textbf{75.0}\stdsub{3.0} & \cellcolor{winrow}\textbf{51.0}\stdsub{1.8} & \cellcolor{winrow}\textbf{40.2}\stdsub{2.2} & \cellcolor{winrow}56.1\stdsub{2.9} & \cellcolor{winrow}\textbf{62.3}\stdsub{1.7} & \cellcolor{winrow}\textbf{62.2}\stdsub{3.2} & \cellcolor{winrow}\textbf{60.1}\dgain{1.6} \\
\midrule

\multirow{7}{*}{\modelTag{\iconLLaVA}{LLaVA-1.6-7B}}
& ProTeGi  & 58.4\stdsub{0.9} & \textbf{72.7}\stdsub{0.1} & \underline{76.8}\stdsub{0.4} & 45.6\stdsub{0.6} & 38.6\stdsub{0.2} & 25.5\stdsub{0.9} & 45.3\stdsub{1.8} & \textbf{81.6}\stdsub{1.2} & 55.6 \\
& OPRO     & 58.3\stdsub{1.6} & 65.6\stdsub{0.3} & 71.9\stdsub{1.2} & 41.1\stdsub{0.8} & 36.3\stdsub{0.2} & 32.3\stdsub{3.7} & 49.5\stdsub{2.4} & 80.1\stdsub{0.5} & 54.4 \\
& DSPy     & 58.5\stdsub{0.6} & 66.9\stdsub{1.1} & 74.5\stdsub{0.9} & 44.3\stdsub{1.7} & 37.8\stdsub{0.7} & 31.4\stdsub{5.2} & 49.4\stdsub{3.0} & 80.3\stdsub{1.3} & 55.4 \\
& TG       & 58.9\stdsub{2.4} & 70.1\stdsub{2.6} & 75.7\stdsub{0.2} & 43.8\stdsub{1.1} & 34.9\stdsub{2.0} & 30.9\stdsub{2.1} & 45.9\stdsub{3.4} & 76.1\stdsub{6.8} & 54.5 \\
& RE       & 56.4\stdsub{1.2} & 69.6\stdsub{2.8} & 76.1\stdsub{0.5} & 45.1\stdsub{0.3} & 37.6\stdsub{1.4} & 29.1\stdsub{1.1} & 43.6\stdsub{1.8} & 80.4\stdsub{0.6} & 54.7 \\
& \cellcolor{secrow}CMVF$^{\!*}$ & \cellcolor{secrow}\textbf{60.3}\stdsub{0.3} & \cellcolor{secrow}70.9\stdsub{1.0} & \cellcolor{secrow}75.8\stdsub{0.9} & \cellcolor{secrow}\underline{48.0}\stdsub{2.5} & \cellcolor{secrow}\underline{40.6}\stdsub{0.2} & \cellcolor{secrow}\underline{37.7}\stdsub{3.7} & \cellcolor{secrow}\textbf{53.5}\stdsub{1.5} & \cellcolor{secrow}81.0\stdsub{0.2} & \cellcolor{secrow}\underline{58.5} \\
& \cellcolor{winrow}\textbf{CMVF} & \cellcolor{winrow}\underline{59.6}\stdsub{0.6} & \cellcolor{winrow}\underline{71.1}\stdsub{0.4} & \cellcolor{winrow}\textbf{77.7}\stdsub{0.6} & \cellcolor{winrow}\textbf{48.2}\stdsub{0.1} & \cellcolor{winrow}\textbf{41.9}\stdsub{3.3} & \cellcolor{winrow}\textbf{38.3}\stdsub{4.4} & \cellcolor{winrow}\underline{51.3}\stdsub{1.9} & \cellcolor{winrow}\underline{81.3}\stdsub{0.4} & \cellcolor{winrow}\textbf{58.7}\dgain{3.1} \\
\midrule

\multirow{7}{*}{\modelTag{\iconPhi}{Phi-3.5-vision}}
& ProTeGi & 58.2\stdsub{2.1} & \underline{72.0}\stdsub{4.9} & 76.0\stdsub{0.6} & 45.5\stdsub{2.4} & 32.5\stdsub{1.6} & 50.9\stdsub{1.8} & 47.4\stdsub{1.7} & 45.8\stdsub{2.0} & 53.5 \\
& OPRO & \underline{59.2}\stdsub{1.0} & 63.8\stdsub{2.5} & 69.5\stdsub{2.2} & 40.4\stdsub{2.9} & 31.9\stdsub{0.8} & 45.0\stdsub{6.4} & 49.3\stdsub{2.0} & 48.1\stdsub{3.9} & 50.9 \\
& DSPy & 57.4\stdsub{1.7} & 64.0\stdsub{1.4} & 74.1\stdsub{2.1} & 42.6\stdsub{1.3} & 34.4\stdsub{1.0} & 51.1\stdsub{1.8} & 51.2\stdsub{1.5} & \textbf{55.7}\stdsub{6.8} & 53.8 \\
& TG & 57.7\stdsub{2.1} & 68.5\stdsub{2.6} & 75.0\stdsub{0.7} & 43.0\stdsub{2.3} & 34.1\stdsub{1.1} & 51.2\stdsub{1.6} & 49.5\stdsub{2.1} & 45.7\stdsub{5.0} & 53.1 \\
& RE & 56.4\stdsub{0.8} & 68.6\stdsub{2.1} & 75.0\stdsub{1.2} & 43.3\stdsub{0.6} & 37.1\stdsub{0.8} & 48.6\stdsub{1.1} & \underline{51.8}\stdsub{0.7} & 48.6\stdsub{3.7} & 53.7 \\
& \cellcolor{secrow}CMVF$^{\!*}$ & \cellcolor{secrow}\textbf{60.1}\stdsub{1.1} & \cellcolor{secrow}71.9\stdsub{0.9} & \cellcolor{secrow}\textbf{77.3}\stdsub{1.6} & \cellcolor{secrow}\underline{46.4}\stdsub{3.0} & \cellcolor{secrow}\textbf{38.5}\stdsub{1.0} & \cellcolor{secrow}\textbf{54.4}\stdsub{2.6} & \cellcolor{secrow}\textbf{52.3}\stdsub{1.4} & \cellcolor{secrow}45.2\stdsub{1.2} & \cellcolor{secrow}\underline{55.8} \\
& \cellcolor{winrow}\textbf{CMVF} & \cellcolor{winrow}58.9\stdsub{1.9} & \cellcolor{winrow}\textbf{72.4}\stdsub{1.5} & \cellcolor{winrow}\underline{77.1}\stdsub{2.3} & \cellcolor{winrow}\textbf{46.5}\stdsub{0.5} & \cellcolor{winrow}\underline{38.0}\stdsub{1.5} & \cellcolor{winrow}\underline{53.4}\stdsub{1.1} & \cellcolor{winrow}51.4\stdsub{1.9} & \cellcolor{winrow}\underline{49.8}\stdsub{1.4} & \cellcolor{winrow}\textbf{55.9}\dgain{2.1} \\
\bottomrule
\end{tabular*}
\caption{Accuracy on 8 benchmarks. Rows group target VLMs; per model the \textbf{CMVF} (our method) row is shaded green and the runner-up by mean gray. \textbf{Bold}/\underline{underline} mark best/second best; gray subscripts show std over 3 runs.}
\label{tab:main_full}
\end{table*}

\paragraph{Headline result.}
\Cref{tab:main_full} restores the full comparison across all five baselines without visual feedback, CMVF, and CMVF$^*$ for all four target VLMs. \textbf{CMVF ranks first on mean accuracy for every target}, with the largest absolute gain on LLaVA-1.6-Mistral-7B (+3.1\,pp) and a comparable +3.0\,pp on Qwen3.5-4B.

\paragraph{Where the wins concentrate.}
The per-dataset pattern is consistent with the visual-feedback hypothesis: gains cluster on visually demanding benchmarks. On Qwen3.5-\allowbreak 4B, CMVF improves RealWorldQA by \textbf{+6.5\,pp}, TextVQA by \textbf{+2.4\,pp}, and PathVQA by \textbf{+1.5\,pp} over the strongest baseline without visual feedback. On LLaVA, ChartQA jumps by \textbf{+6.0\,pp} and PathVQA by \textbf{+3.3\,pp}. On the four additional benchmarks, CMVF again ranks first, by $+1.12$ to $+3.88$\,pp over the strongest baseline (\Cref{tab:official_rescored}). This matches the modeling intuition in \Cref{eq:intuition}: the gain scales with $\alpha_V$, the fraction of visually-grounded errors, which is largest precisely on perception-heavy tasks.

\paragraph{The gains are statistically significant.}
Exact McNemar tests against the strongest baseline, on paired per-example predictions, confirm the improvements: $p=7.9\times10^{-14}$ on LLaVA/TextVQA, $p=0.021$ on LLaVA/ChartQA, and $p=0.015$ on Qwen2.5-VL/TextVQA---significant precisely on OCR, chart-grounding, and fine-grained visual tasks, where visual evidence is essential.

\begin{table}[t]
\centering
\small
\setlength{\tabcolsep}{3pt}
\renewcommand{\arraystretch}{1.05}
\begin{tabular*}{\columnwidth}{@{\extracolsep{\fill}} l ccccc}
\toprule
\textbf{Method} & \textbf{AI2D} & \textbf{DocVQA} & \textbf{MMMU} & \textbf{SciQA} & \textbf{Mean} \\
\midrule
ProTeGi & 78.46 & 68.31 & \underline{47.62} & 81.42 & 68.95 \\
OPRO & 76.83 & 64.28 & 45.89 & \underline{85.30} & 68.08 \\
DSPy & 74.21 & 66.74 & 43.47 & 78.95 & 65.84 \\
TG & 72.68 & 62.53 & 41.16 & 76.63 & 63.25 \\
\rowcolor{secrow}
RE & \underline{80.91} & \underline{70.86} & \underline{47.62} & 83.07 & \underline{70.62} \\
\rowcolor{winrow}
\textbf{CMVF} & \textbf{84.62} & \textbf{74.74} & \textbf{48.74} & \textbf{87.60} & \textbf{73.93} \\
\bottomrule
\end{tabular*}
\caption{Accuracy (\%) on the four additional benchmarks (Qwen3.5-4B). The \textbf{CMVF} row is highlighted; the runner-up by mean (RE) is shaded gray. \textbf{Bold}: best; \underline{underline}: best baseline (ProTeGi and RE tie on MMMU). CMVF ranks first on all four and on their mean.}
\label{tab:official_rescored}
\end{table}

\paragraph{It is not the rewriter.}
CMVF$^*$ uses a REVOLVE-style rewriter on top of the same visual-gradient pipeline. Relative to its counterpart without visual feedback, adding the visual-gradient channel improves a REVOLVE-style update on most targets (e.g., 70.3$\rightarrow$71.9 on Qwen3.5-4B, 54.7$\rightarrow$58.5 on LLaVA, and 53.7$\rightarrow$55.8 on Phi). This shows that the visual channel is not tied to a TextGrad-style rewriter, while the remaining gap to CMVF also makes clear that the rewriter choice still matters.

\subsection{A2: The Optimizer Self-Discovers Expert-Style Visual Inspection Checklists}
\label{sec:emergent}

Beyond the headline accuracy gains, we observe a striking qualitative effect: \textbf{the optimizer VLM is not given dataset-specific visual failure rules, yet it converges on instructions that look like a domain expert's checklist}. This emergent self-diagnostic behavior is, in our view, the most interesting observation of this paper.

\paragraph{Prompt contrast and source audit.}
\Cref{tab:cmvf_trace} lists representative deployed prompts on Qwen3.5-\allowbreak 4B. Baselines without visual feedback converge on \emph{generic, vision-agnostic} hedges (``read step by step,'' ``be precise''), while CMVF produces \emph{task-specific visual checklists}. Crucially, this is not an artifact of hand-written rules: the Stage-1 prompt is deliberately broad (``key diagnostic finding'' for medical, ``key visual details'' for general) and never mentions chart axes, small-text OCR, slice-level checks, or local-object priors---yet the final prompts recover exactly these task-level routines from failed-image evidence.

\begin{table*}[!t]
\centering
\scriptsize
\setlength{\tabcolsep}{3.5pt}
\renewcommand{\arraystretch}{1.05}
\begin{tabular}{l p{14.2cm} c}
\toprule
\textbf{Method} & \textbf{Deployed prompt (visual-specific clauses highlighted)} & \textbf{Acc} \\
\midrule
\rowcolor{rowmed}
\multicolumn{3}{l}{\textcolor{badgemed}{$\blacksquare$}~\textbf{TextVQA}\;\textit{\scriptsize(scene text, OCR)}} \\
\rowcolor{rowmed}
\meth{mOPRO}{OPRO}
 & \weakp{Imagine your mind is a vast, still pond where ripples of distraction fade away until only the single drop of truth answering the question can be seen; gently surface to inspect this one isolated ripple\ldots}
 & 69.1 \\
\rowcolor{rowmed}
\meth{mTG}{TG}
 & \weakp{Analyze the image to extract only visual information, \textbf{explicitly ignoring any text}, labels, or metadata that cannot be verified by visual inspection alone.}
 & 63.8 \\
\rowcolor{rowmed}
\methcmvf
 & \addt{Meticulously scan the entire image to \textbf{verify specific counts}, \textbf{read small text}, and confirm fine details rather than relying on generic associations or prominent but unrelated features.}
 & \textbf{71.5}\,\textcolor{gainred}{\textbf{$\uparrow$2.4}} \\
\midrule
\rowcolor{rowchart}
\multicolumn{3}{l}{\textcolor{badgechart}{$\blacksquare$}~\textbf{ChartQA}\;\textit{\scriptsize(chart reasoning, axes/legends)}} \\
\rowcolor{rowchart}
\meth{mTG}{TG}
 & \weakp{Examine the chart and answer the question. Be precise and double-check your interpretation before giving the final answer.}
 & 78.6 \\
\rowcolor{rowchart}
\meth{mRE}{RE}
 & \weakp{Analyze the chart step by step, revise any inconsistent interpretation, and return the final value or category.}
 & 75.1 \\
\rowcolor{rowchart}
\methcmvf
 & \addt{\textbf{Read axis labels, units, and the legend mapping} before reasoning about magnitudes; identify the requested data point and \textbf{verify its exact value} before answering.}
 & \textbf{81.4}\,\textcolor{gainred}{\textbf{$\uparrow$2.8}} \\
\midrule
\rowcolor{rowreal}
\multicolumn{3}{l}{\textcolor{badgereal}{$\blacksquare$}~\textbf{RealWorldQA}\;\textit{\scriptsize(real-world VQA, scene reasoning)}} \\
\rowcolor{rowreal}
\meth{mDSPy}{DSPy}
 & \weakp{You are a visual question answering assistant. Inspect the image and answer the question accurately.}
 & 82.6 \\
\rowcolor{rowreal}
\meth{mTG}{TG}
 & \weakp{Look at the image carefully and answer the question. Make sure your reasoning is grounded in what is actually visible.}
 & \underline{84.8} \\
\rowcolor{rowreal}
\methcmvf
 & \addt{\textbf{Scan for small, low-contrast, occluded, or background objects}; verify orientation, signs, and object states; \textbf{avoid global scene priors} when the question is about a local object.}
 & \textbf{91.3}\,\textcolor{gainred}{\textbf{$\uparrow$6.5}} \\
\bottomrule
\end{tabular}
\caption{Representative deployed prompts on Qwen3.5-4B. For each benchmark we show two baselines (without visual feedback, gray italics) and CMVF (green). \textbf{Bold} marks the decisive clause: baselines stay generic or, on TextVQA, even instruct the model to \emph{ignore} text, whereas CMVF names the exact visual check the task demands. Acc is test accuracy.}
\label{tab:cmvf_trace}
\end{table*}

\paragraph{What CMVF rewrites the prompt to say.}
Feedback without visual inspection typically yields generic advice (``verify your reasoning against the visual evidence,'' ``be careful with details''). CMVF instead discovers \emph{task-specific} routines across the 8 benchmarks: \textbf{(1)} SLAKE-CT/MRI prompts ask the model to verify anatomical slice level and separate similar-density regions; \textbf{(2)} VQA-RAD/PathVQA prompts locate the focal abnormality before classification; \textbf{(3)} ChartQA prompts read axes, units, and legends before comparing values; \textbf{(4)} TextVQA prompts scan for small or non-salient text; and \textbf{(5)} RealWorldQA prompts check occlusion and avoid global scene priors for local-object questions. These instructions are not pre-specified in the seed prompt or Stage-1 description prompt; they arise from Stage-2 aggregation over Stage-1 perception, showing that a frozen optimizer VLM can self-organize reusable visual checklists from failed-image evidence alone.

\paragraph{Why this matters.}
The phenomenon suggests that the bottleneck of multimodal APO is not optimization machinery but \emph{evidence}: once the optimizer can see failed images, recurring visual failure modes become extractable in language and can be carried by a static inference-time text prompt.

\subsection{A3: Optimized Strategies Transfer Across Target Models}
\label{sec:transfer}

If CMVF merely memorized target-specific quirks, its prompts would not survive being moved to a different model. We test this directly: the CMVF prompt optimized on each of Qwen2.5-VL-7B, LLaVA-1.6-7B, and Phi-3.5-vision is deployed \emph{unchanged} on Qwen3.5-4B, with \emph{no} re-optimization (\Cref{tab:transfer}). Every transferred prompt outperforms a TextGrad prompt optimized \emph{directly} on Qwen3.5-4B (70.3), by $+1.2$ to $+2.8$\,pp on the 8-benchmark mean, and the advantage holds per-dataset rather than on a few benchmarks---on ChartQA a transferred prompt even exceeds the in-domain one. This indicates that Stage-2 aggregation surfaces reusable, \emph{task-level} visual strategies rather than spurious correlations bound to a single model, directly addressing the concern that access to ground-truth labels could induce dataset- or target-specific overfitting. The in-domain prompt optimized on 4B remains best overall (75.0), as expected.

\begin{table*}[t]
\centering
\small
\renewcommand{\arraystretch}{1.05}
\begin{tabular*}{\textwidth}{@{\extracolsep{\fill}} l cccccccc c}
\toprule
\textbf{Prompt source $\to$ Qwen3.5-4B} & \textbf{CT} & \textbf{MRI} & \textbf{X-Ray} & \textbf{RAD} & \textbf{Path} & \textbf{Chart} & \textbf{Text} & \textbf{RWQA} & \textbf{Mean} \\
\midrule
TextGrad (direct on 4B) & 75.9 & 75.3 & 80.7 & 57.9 & 45.4 & 78.6 & 63.8 & 84.8 & 70.3 \\
CMVF $\leftarrow$ Qwen2.5-VL-7B & 76.0 & 76.8 & 82.5 & 57.4 & 44.8 & 83.2 & 66.1 & 84.8 & 71.5 \\
CMVF $\leftarrow$ LLaVA-1.6-7B & 76.4 & 78.7 & 83.4 & 59.4 & 48.2 & 80.5 & 65.5 & 84.3 & 72.1 \\
\rowcolor{secrow}
CMVF $\leftarrow$ Phi-3.5-vision & 76.0 & 79.0 & 83.0 & 58.3 & 47.6 & \textbf{84.3} & 69.7 & 87.0 & 73.1 \\
\rowcolor{winrow}
CMVF (direct on 4B) & \textbf{77.9} & \textbf{80.7} & \textbf{84.5} & \textbf{63.4} & \textbf{49.0} & 81.4 & \textbf{71.5} & \textbf{91.3} & \textbf{75.0} \\
\bottomrule
\end{tabular*}
\caption{Cross-target transfer: a CMVF prompt optimized on one model and deployed on Qwen3.5-4B \emph{without} re-optimization still beats TextGrad optimized directly on 4B. Every transferred prompt exceeds the direct TextGrad baseline; on ChartQA transfer even beats the in-domain prompt. Bold: best per column.}
\label{tab:transfer}
\end{table*}

\subsection{A4: Image-Aware Perception Drives the Gains, Not Extra Tokens}
\label{sec:ablation}

\begin{table}[t]
\centering
\small
\setlength{\tabcolsep}{3pt}
\begin{tabular*}{\columnwidth}{@{\extracolsep{\fill}} l c cc cc}
\toprule
& \textbf{Qwen3.5-4B} & \multicolumn{2}{c}{\textbf{Qwen2.5-VL-7B}} & \multicolumn{2}{c}{\textbf{LLaVA-1.6-7B}} \\
\cmidrule(lr){2-2}\cmidrule(lr){3-4}\cmidrule(lr){5-6}
\textbf{Method} & \textbf{Mean} & \textbf{RWQA} & \textbf{Text} & \textbf{RWQA} & \textbf{Text} \\
\midrule
CM-Blind & 71.9 & 56.1 & 56.1 & 69.1 & 39.5 \\
CM-Caption & 72.2 & 58.3 & 58.2 & 71.3 & 41.5 \\
\rowcolor{secrow} CM-Vis & 72.8 & 60.5 & 60.4 & 73.5 & 43.8 \\
\rowcolor{winrow} \textbf{CMVF} & \textbf{75.0} & \textbf{62.2} & \textbf{62.3} & \textbf{81.3} & \textbf{51.3} \\
\bottomrule
\end{tabular*}
\caption{Feedback-channel ablation across three target VLMs (methods defined in text; token-budget matched). \emph{Qwen3.5-4B} is the 8-benchmark mean; the other two targets are per-dataset (official protocol). Every column rises monotonically---CMVF (green) best, CM-Vis (gray) runner-up.}
\label{tab:ablation_compact}
\end{table}

\begin{figure*}[t]
\centering
\includegraphics[width=0.94\textwidth]{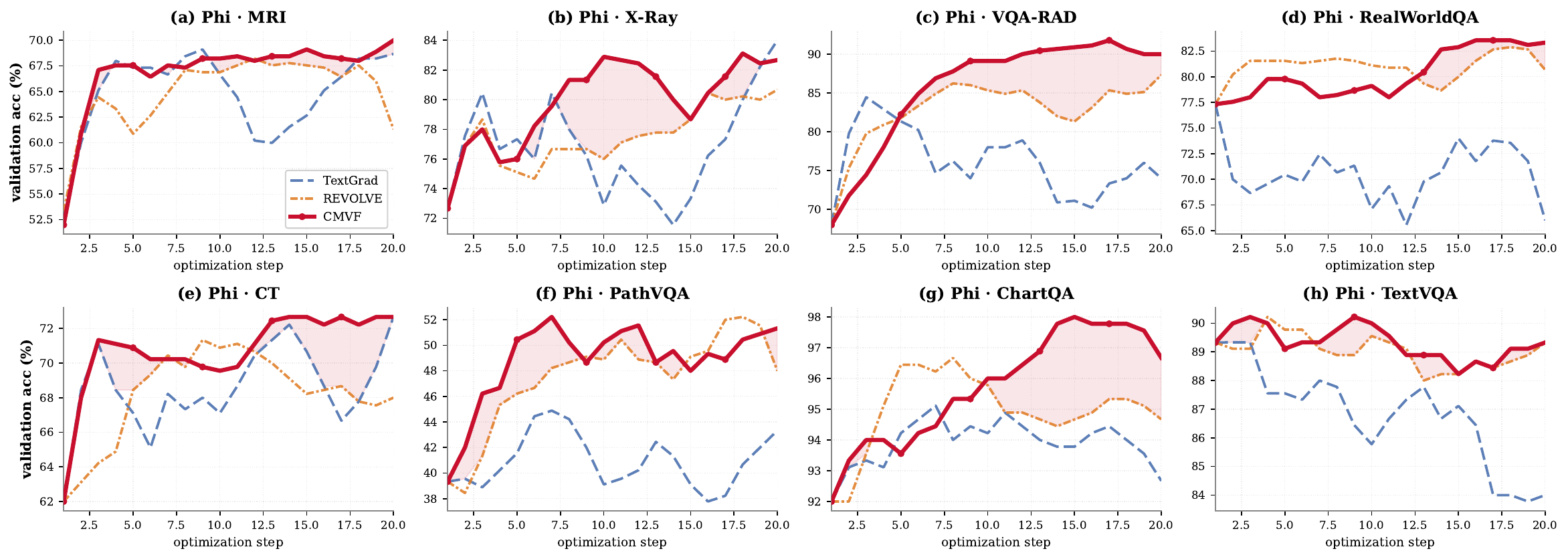}
\caption{Validation accuracy over 20 optimization steps on eight Phi-3.5-vision benchmarks (top: MRI, X-Ray, VQA-RAD, RealWorldQA; bottom: CT, PathVQA, ChartQA, TextVQA). CMVF (red) stays above both text-only baselines (TextGrad, REVOLVE) and keeps improving after they plateau; the shaded band marks CMVF's margin over the stronger baseline.}
\label{fig:train_curves}
\end{figure*}

\paragraph{Tokens are cheap; pixels are decisive.}
A2 attributed the gains to image-aware evidence; \Cref{tab:ablation_compact} now rules out longer feedback as the cause under a token-budget-matched protocol on Qwen3.5-4B. \textbf{CM-Blind} adds the same number of feedback tokens as CMVF but \emph{without image access}; it reaches only 71.9 (\textbf{+1.6\,pp} over TG). \textbf{CM-Caption} feeds the optimizer generic, question-agnostic captions and reaches 72.2. \textbf{CM-Vis} keeps the visual channel but drops the fused error attribution---concatenating raw per-image descriptions into the rewrite prompt, closely matching the IPO-style~\cite{du2024ipo} recipe adapted from CLIP class-template prompts to VQA instruction prompts---reaching 72.8, while the full \textbf{CMVF} adds Stage-2 aggregation over \emph{all} wrong samples and reaches \textbf{75.0}. The implication is direct: longer prompts alone account for at most $+1.6$\,pp over TG, while \emph{question-aware visual perception with aggregation} accounts for the remaining $+3.1$\,pp. CMVF's \emph{deployed} prompt averages $\sim$205 BPE tokens (excluding \texttt{\{fs\}}), comparable to TG/RE (185--192) and 14--26\% shorter than DSPy/OPRO/ProTeGi (239--276); the gain cannot be attributed to a longer inference-time prompt.

\paragraph{The ordering generalizes across targets.}
The same monotone channel progression---CM-Blind $<$ CM-Caption $<$ CM-Vis $<$ CMVF---holds on \emph{every} (target, dataset) pair for two additional target VLMs (\Cref{tab:ablation_compact}), so the effect is not specific to Qwen3.5-4B. The margin widens on the cross-family target LLaVA-1.6-7B (CMVF $+7.8$ over CM-Vis on RealWorldQA and $+7.5$ on TextVQA), consistent with a weaker base model leaving more visually-groundable headroom for the visual channel to recover.

Two alternative Stage-2 aggregation modes further isolate \emph{why} unified aggregation is preferred: \textbf{CM-Mem} ($72.1$, $-2.9$ vs.\ CMVF), which carries a cross-step memory pool and is misled by stale signals from earlier weaker prompts, and \textbf{Diff} ($72.2$, $-2.8$), which fragments the visual signal by partitioning errors before aggregation. Both underperform unified CMVF (full design-space sweep in the appendix), confirming that the gain comes from question-aware perception of \emph{all} wrong samples with a single unified aggregation step, not from any individual component.

\subsection{A5: CMVF Generalizes Better at \texttt{\textasciitilde}25\% Training-Time Overhead and No Additional Inference Cost}
\label{sec:efficiency_generalization}

\begin{figure}[t]
\centering
\includegraphics[width=0.84\columnwidth]{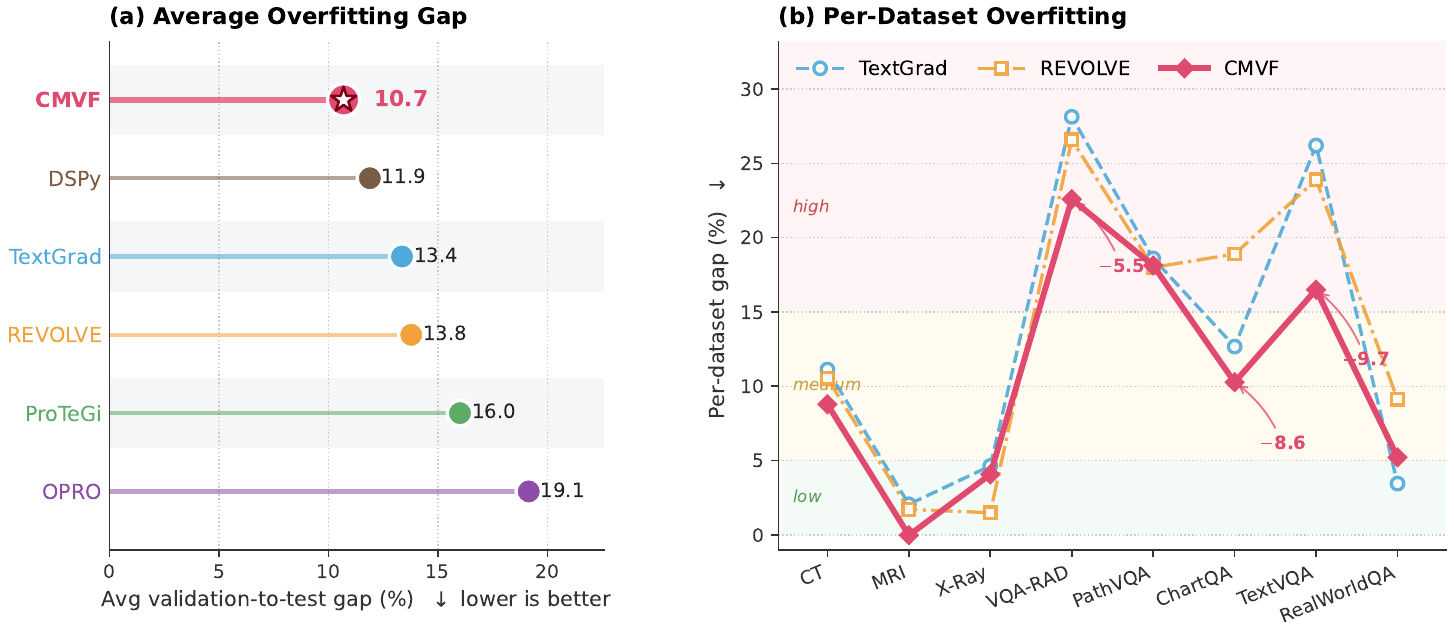}
\caption{Generalization gap (validation-best accuracy minus test accuracy; lower is better) on 8 benchmarks. CMVF has the smallest average gap and reduces high-gap cases such as ChartQA and VQA-RAD.}
\label{fig:gen_gap}
\end{figure}

\paragraph{Late-stage improvement.}
\Cref{fig:train_curves} shows that CMVF \emph{continues} improving after baselines without visual feedback flatten on these settings. Once feedback without visual inspection has corrected the linguistically obvious failures, the remaining errors are more likely to be visually grounded, so an image-grounded gradient can keep moving.

\paragraph{Generalization gap.}
\Cref{fig:gen_gap}(a) shows that CMVF has the \emph{smallest} average validation-to-test gap among all six methods, and \Cref{fig:gen_gap}(b) confirms this advantage holds per-dataset rather than coming from a few easy benchmarks. We attribute this to aggregation: instead of overfitting prompts to isolated validation errors, CMVF rewrites them around \emph{recurring} visual patterns---a form of implicit regularization.

\paragraph{Cost in practice.}
The visual-gradient pipeline adds a one-shot training-time overhead but \emph{no additional} inference-time cost. End-to-end optimization on Qwen3.5-4B takes \textbf{2.6\,h} for CMVF versus 2.1\,h for TextGrad and 2.4\,h for REVOLVE---a $\sim\!20$--$25\%$ wall-clock increase that buys $+3.0$\,pp mean accuracy. Compared to ProTeGi (5.9\,h), CMVF is more than $2{\times}$ faster while delivering substantially higher accuracy. Crucially, the final deployed artifact is a standard text prompt, so inference uses the \emph{same serving interface and call count} as a vanilla deployment.

\section{Related Work}
\label{sec:related}

\paragraph{Automatic prompt optimization.}
Early APO methods search over prompt candidates or use LLMs as meta-optimizers~\cite{yang2024large,zhou2022large}, including reinforcement-learning, black-box, and evolutionary variants~\cite{deng2022rlprompt,chen2023instructzero,fernando2023promptbreeder,guo2024connecting}. DSPy compiles declarative LLM programs into optimized prompt pipelines~\cite{khattab2024dspy}, gradient-inspired methods such as ProTeGi and TextGrad use natural-language critiques as textual gradients~\cite{pryzant2023automatic,yuksekgonul2025optimizing}, and REVOLVE tracks response evolution across iterations~\cite{zhang2024revolve}. As prompts are highly sensitive to surface form~\cite{sclar2024quantifying}, recent work also explores symbolic and structure-factorized prompt spaces~\cite{ou2025symbolic,liu2026adaptivepromptstructurefactorization}. These methods keep the feedback channel textual even when the downstream task is multimodal.

\paragraph{Prompt learning for vision-language models.}
A parallel line adapts VLMs by learning \emph{continuous} prompts: CoOp and CoCoOp tune soft context vectors for CLIP-style classifiers~\cite{zhou2022coop,zhou2022conditional}, VPT prepends learnable visual tokens~\cite{jia2022visual}, and MaPLe couples prompts across the vision and language branches~\cite{khattak2023maple}. These require gradient access to the backbone and deploy a non-textual prompt, whereas CMVF optimizes a \emph{discrete, human-readable} instruction and treats the target VLM as a black box at both optimization and deployment time.

\paragraph{Multimodal prompting, visual feedback, and self-refinement.}
The closest prior work is \textbf{IPO}~\cite{du2024ipo}, which lets an LMM describe training images and rewrites a CLIP class-template prompt for zero-shot classification. CMVF shares its recipe of \emph{answer-agnostic visual descriptions + LMM optimizer + textual-prompt deployment}, but inspects only the wrong set $\mathcal{W}_t$, adds a Stage-2 error-aware aggregation into reusable blind-spot patterns (removing it costs 2.2\,pp; \Cref{tab:ablation_compact}), and targets generative VQA prompts on four VLMs (4B--7B) rather than a fixed CLIP backbone. Concurrently, \emph{multimodal prompt optimization} (MPO)~\cite{choi2025multimodal} keeps the deployed prompt multimodal at a per-query visual cost, whereas CMVF deploys a standard textual prompt. Unlike self-refinement methods that critique outputs per instance at inference time~\cite{madaan2023self,shinn2023reflexion}, CMVF aggregates cross-example visual evidence \emph{once}, at optimization time.

\section{Conclusion}
\label{sec:conclusion}

We introduce CMVF, a two-stage method that lets the optimizer VLM inspect the images the target model failed on and compiles that evidence into an ordinary text prompt. CMVF ranks first on all four target VLMs, beating the strongest baseline by $+2.4$\,pp on average (up to $+6.5$\,pp) at no added inference cost, and the optimizer self-organizes into transferable expert-style visual checklists. The bottleneck of multimodal APO is \emph{visual evidence}, not optimization machinery.

\bibliography{refs}

\begin{thebibliography}{34}
\providecommand{\natexlab}[1]{#1}

\bibitem[{Abdin et~al.(2024)Abdin, Aneja, Awadalla, Awadallah, Awan, Bach, Bahree et~al.}]{abdin2024phi3}
Abdin, M.; Aneja, J.; Awadalla, H.; Awadallah, A.; Awan, A.~A.; Bach, N.; Bahree, A.; et~al. 2024.
\newblock Phi-3 Technical Report: A Highly Capable Language Model Locally on Your Phone.
\newblock \emph{arXiv preprint arXiv:2404.14219}.

\bibitem[{Chen et~al.(2024)Chen, Chen, Goldstein, Huang, and Zhou}]{chen2023instructzero}
Chen, L.; Chen, J.; Goldstein, T.; Huang, H.; and Zhou, T. 2024.
\newblock {InstructZero}: Efficient Instruction Optimization for Black-Box Large Language Models.
\newblock In \emph{Proceedings of the 41st International Conference on Machine Learning}, volume 235 of \emph{Proceedings of Machine Learning Research}, 6503--6518. PMLR.

\bibitem[{Choi et~al.(2026)Choi, Kim, Baek, and Hwang}]{choi2025multimodal}
Choi, Y.; Kim, D.; Baek, J.; and Hwang, S.~J. 2026.
\newblock Multimodal Prompt Optimization: Why Not Leverage Multiple Modalities for MLLMs.
\newblock In \emph{International Conference on Learning Representations}.

\bibitem[{Deng et~al.(2022)Deng, Wang, Hsieh, Wang, Guo, Shu, Song, Xing, and Hu}]{deng2022rlprompt}
Deng, M.; Wang, J.; Hsieh, C.-P.; Wang, Y.; Guo, H.; Shu, T.; Song, M.; Xing, E.~P.; and Hu, Z. 2022.
\newblock {RLPrompt}: Optimizing Discrete Text Prompts with Reinforcement Learning.
\newblock In \emph{Proceedings of the 2022 Conference on Empirical Methods in Natural Language Processing (EMNLP)}, 3369--3391.

\bibitem[{Du, Sun, and Snoek(2024)}]{du2024ipo}
Du, Y.; Sun, W.; and Snoek, C.~G. 2024.
\newblock {IPO}: Interpretable Prompt Optimization for Vision-Language Models.
\newblock \emph{Advances in Neural Information Processing Systems}, 37: 126725--126766.

\bibitem[{Fernando et~al.(2024)Fernando, Banarse, Michalewski, Osindero, and Rockt{\"a}schel}]{fernando2023promptbreeder}
Fernando, C.; Banarse, D.~S.; Michalewski, H.; Osindero, S.; and Rockt{\"a}schel, T. 2024.
\newblock {Promptbreeder}: Self-Referential Self-Improvement via Prompt Evolution.
\newblock In \emph{Proceedings of the 41st International Conference on Machine Learning}, volume 235 of \emph{Proceedings of Machine Learning Research}, 13481--13544. PMLR.

\bibitem[{Guo et~al.(2024)Guo, Wang, Guo, Li, Song, Tan, Liu, Bian, and Yang}]{guo2024connecting}
Guo, Q.; Wang, R.; Guo, J.; Li, B.; Song, K.; Tan, X.; Liu, G.; Bian, J.; and Yang, Y. 2024.
\newblock Connecting Large Language Models with Evolutionary Algorithms Yields Powerful Prompt Optimizers.
\newblock In \emph{International Conference on Learning Representations}.

\bibitem[{He et~al.(2020)He, Zhang, Mou, Xing, and Xie}]{he2020pathvqa}
He, X.; Zhang, Y.; Mou, L.; Xing, E.; and Xie, P. 2020.
\newblock {PathVQA}: 30000+ Questions for Medical Visual Question Answering.
\newblock \emph{arXiv preprint arXiv:2003.10286}.

\bibitem[{Jia et~al.(2022)Jia, Tang, Chen, Cardie, Belongie, Hariharan, and Lim}]{jia2022visual}
Jia, M.; Tang, L.; Chen, B.-C.; Cardie, C.; Belongie, S.; Hariharan, B.; and Lim, S.-N. 2022.
\newblock Visual prompt tuning.
\newblock In \emph{European Conference on Computer Vision}, 709--727. Springer.

\bibitem[{Kembhavi et~al.(2016)Kembhavi, Salvato, Kolve, Seo, Hajishirzi, and Farhadi}]{kembhavi2016diagram}
Kembhavi, A.; Salvato, M.; Kolve, E.; Seo, M.; Hajishirzi, H.; and Farhadi, A. 2016.
\newblock A diagram is worth a dozen images.
\newblock In \emph{European Conference on Computer Vision}, 235--251. Springer.

\bibitem[{Khattab et~al.(2024)Khattab, Singhvi, Maheshwari, Zhang, Santhanam, Vardhamanan, Haq, Sharma, Joshi, Moazam, Miller, Zaharia, and Potts}]{khattab2024dspy}
Khattab, O.; Singhvi, A.; Maheshwari, P.; Zhang, Z.; Santhanam, K.; Vardhamanan, S.; Haq, S.; Sharma, A.; Joshi, T.~T.; Moazam, H.; Miller, H.; Zaharia, M.; and Potts, C. 2024.
\newblock {DSPy}: Compiling Declarative Language Model Calls into State-of-the-Art Pipelines.
\newblock In \emph{International Conference on Learning Representations}.

\bibitem[{Khattak et~al.(2023)Khattak, Rasheed, Maaz, Khan, and Khan}]{khattak2023maple}
Khattak, M.~U.; Rasheed, H.; Maaz, M.; Khan, S.; and Khan, F.~S. 2023.
\newblock {MaPLe}: Multi-Modal Prompt Learning.
\newblock In \emph{Proceedings of the IEEE/CVF Conference on Computer Vision and Pattern Recognition}, 19113--19122.

\bibitem[{Lau et~al.(2018)Lau, Gayen, Ben~Abacha, and Demner-Fushman}]{lau2018dataset}
Lau, J.~J.; Gayen, S.; Ben~Abacha, A.; and Demner-Fushman, D. 2018.
\newblock A dataset of clinically generated visual questions and answers about radiology images.
\newblock \emph{Scientific Data}, 5(1): 180251.

\bibitem[{Li et~al.(2023)Li, Wong, Zhang, Usuyama, Liu, Yang, Naumann, Poon, and Gao}]{li2023llava}
Li, C.; Wong, C.; Zhang, S.; Usuyama, N.; Liu, H.; Yang, J.; Naumann, T.; Poon, H.; and Gao, J. 2023.
\newblock {LLaVA-Med}: Training a Large Language-and-Vision Assistant for Biomedicine in One Day.
\newblock \emph{Advances in Neural Information Processing Systems}, 36: 28541--28564.

\bibitem[{Liu et~al.(2021)Liu, Zhan, Xu, Ma, Yang, and Wu}]{liu2021slake}
Liu, B.; Zhan, L.-M.; Xu, L.; Ma, L.; Yang, Y.; and Wu, X.-M. 2021.
\newblock {SLAKE}: A Semantically-Labeled Knowledge-Enhanced Dataset for Medical Visual Question Answering.
\newblock In \emph{2021 IEEE 18th International Symposium on Biomedical Imaging (ISBI)}, 1650--1654. IEEE.

\bibitem[{Liu et~al.(2026)Liu, Wang, Guo, Shou, and Tang}]{liu2026adaptivepromptstructurefactorization}
Liu, H.; Wang, Z.; Guo, Y.; Shou, H.; and Tang, X. 2026.
\newblock Adaptive Prompt Structure Factorization: A Framework for Self-Discovering and Optimizing Compositional Prompt Programs.
\newblock In \emph{Proceedings of the 64th Annual Meeting of the Association for Computational Linguistics (Volume 1: Long Papers)}, 11690--11714. Association for Computational Linguistics.

\bibitem[{Lu et~al.(2022)Lu, Mishra, Xia, Qiu, Chang, Zhu, Tafjord, Clark, and Kalyan}]{lu2022learn}
Lu, P.; Mishra, S.; Xia, T.; Qiu, L.; Chang, K.-W.; Zhu, S.-C.; Tafjord, O.; Clark, P.; and Kalyan, A. 2022.
\newblock Learn to explain: Multimodal reasoning via thought chains for science question answering.
\newblock \emph{Advances in Neural Information Processing Systems}, 35: 2507--2521.

\bibitem[{Madaan et~al.(2023)Madaan, Tandon, Gupta, Hallinan, Gao, Wiegreffe, Alon, Dziri, Prabhumoye, Yang, Gupta, Majumder, Hermann, Welleck, Yazdanbakhsh, and Clark}]{madaan2023self}
Madaan, A.; Tandon, N.; Gupta, P.; Hallinan, S.; Gao, L.; Wiegreffe, S.; Alon, U.; Dziri, N.; Prabhumoye, S.; Yang, Y.; Gupta, S.; Majumder, B.~P.; Hermann, K.; Welleck, S.; Yazdanbakhsh, A.; and Clark, P. 2023.
\newblock Self-Refine: Iterative Refinement with Self-Feedback.
\newblock In \emph{Advances in Neural Information Processing Systems (NeurIPS)}, volume~36, 46534--46594.

\bibitem[{Masry et~al.(2022)Masry, Long, Tan, Joty, and Hoque}]{masry2022chartqa}
Masry, A.; Long, D.~X.; Tan, J.~Q.; Joty, S.; and Hoque, E. 2022.
\newblock {ChartQA}: A Benchmark for Question Answering about Charts with Visual and Logical Reasoning.
\newblock In \emph{Findings of the Association for Computational Linguistics: ACL 2022}, 2263--2279.

\bibitem[{Mathew, Karatzas, and Jawahar(2021)}]{mathew2021docvqa}
Mathew, M.; Karatzas, D.; and Jawahar, C. 2021.
\newblock {DocVQA}: A Dataset for {VQA} on Document Images.
\newblock In \emph{Proceedings of the IEEE/CVF Winter Conference on Applications of Computer Vision}, 2200--2209.

\bibitem[{Ou et~al.(2025)Ou, Zhou, Ding, Li, Wu, Wang, Chen, Wang, Xu, Zhang, Chen, and Jiang}]{ou2025symbolic}
Ou, Y.; Zhou, W.; Ding, S.; Li, L.; Wu, J.; Wang, T.; Chen, J.; Wang, S.; Xu, X.; Zhang, N.; Chen, H.; and Jiang, Y.~E. 2025.
\newblock Symbolic learning enables self-evolving agents.
\newblock \emph{AI Open}, 6: 314--322.

\bibitem[{Pryzant et~al.(2023)Pryzant, Iter, Li, Lee, Zhu, and Zeng}]{pryzant2023automatic}
Pryzant, R.; Iter, D.; Li, J.; Lee, Y.; Zhu, C.; and Zeng, M. 2023.
\newblock Automatic Prompt Optimization with ``Gradient Descent'' and Beam Search.
\newblock In \emph{Proceedings of the 2023 Conference on Empirical Methods in Natural Language Processing}, 7957--7968.

\bibitem[{Sclar et~al.(2024)Sclar, Choi, Tsvetkov, and Suhr}]{sclar2024quantifying}
Sclar, M.; Choi, Y.; Tsvetkov, Y.; and Suhr, A. 2024.
\newblock Quantifying Language Models' Sensitivity to Spurious Features in Prompt Design or: How I learned to start worrying about prompt formatting.
\newblock In \emph{International Conference on Learning Representations}.

\bibitem[{Shinn et~al.(2023)Shinn, Cassano, Gopinath, Narasimhan, and Yao}]{shinn2023reflexion}
Shinn, N.; Cassano, F.; Gopinath, A.; Narasimhan, K.; and Yao, S. 2023.
\newblock Reflexion: Language Agents with Verbal Reinforcement Learning.
\newblock In \emph{Advances in Neural Information Processing Systems (NeurIPS)}, volume~36, 8634--8652.

\bibitem[{Singh et~al.(2019)Singh, Natarajan, Shah, Jiang, Chen, Batra, Parikh, and Rohrbach}]{singh2019towards}
Singh, A.; Natarajan, V.; Shah, M.; Jiang, Y.; Chen, X.; Batra, D.; Parikh, D.; and Rohrbach, M. 2019.
\newblock Towards {VQA} Models That Can Read.
\newblock In \emph{Proceedings of the IEEE/CVF Conference on Computer Vision and Pattern Recognition}, 8317--8326.

\bibitem[{Wang et~al.(2024)Wang, Bai, Tan, Wang, Fan, Bai, Chen, Liu, Wang, Ge et~al.}]{wang2024qwen2vl}
Wang, P.; Bai, S.; Tan, S.; Wang, S.; Fan, Z.; Bai, J.; Chen, K.; Liu, X.; Wang, J.; Ge, W.; et~al. 2024.
\newblock {Qwen2-VL}: Enhancing Vision-Language Model's Perception of the World at Any Resolution.
\newblock \emph{arXiv preprint arXiv:2409.12191}.

\bibitem[{{xAI}(2024)}]{xai2024realworldqa}
{xAI}. 2024.
\newblock {RealWorldQA}.
\newblock \url{https://huggingface.co/datasets/xai-org/RealworldQA}.
\newblock Dataset on Hugging Face; accessed 2026-07-24.

\bibitem[{Yang et~al.(2024)Yang, Wang, Lu, Liu, Le, Zhou, and Chen}]{yang2024large}
Yang, C.; Wang, X.; Lu, Y.; Liu, H.; Le, Q.~V.; Zhou, D.; and Chen, X. 2024.
\newblock Large language models as optimizers.
\newblock In \emph{International Conference on Learning Representations}.

\bibitem[{Yue et~al.(2024)Yue, Ni, Zhang, Zheng, Liu, Zhang, Stevens, Jiang, Ren, Sun, Wei, Yu, Yuan, Sun, Yin, Zheng, Yang, Liu, Huang, Sun, Su, and Chen}]{yue2024mmmu}
Yue, X.; Ni, Y.; Zhang, K.; Zheng, T.; Liu, R.; Zhang, G.; Stevens, S.; Jiang, D.; Ren, W.; Sun, Y.; Wei, C.; Yu, B.; Yuan, R.; Sun, R.; Yin, M.; Zheng, B.; Yang, Z.; Liu, Y.; Huang, W.; Sun, H.; Su, Y.; and Chen, W. 2024.
\newblock {MMMU}: A Massive Multi-Discipline Multimodal Understanding and Reasoning Benchmark for Expert {AGI}.
\newblock In \emph{Proceedings of the IEEE/CVF Conference on Computer Vision and Pattern Recognition}, 9556--9567.

\bibitem[{Yuksekgonul et~al.(2025)Yuksekgonul, Bianchi, Boen, Liu, Lu, Huang, Guestrin, and Zou}]{yuksekgonul2025optimizing}
Yuksekgonul, M.; Bianchi, F.; Boen, J.; Liu, S.; Lu, P.; Huang, Z.; Guestrin, C.; and Zou, J. 2025.
\newblock Optimizing Generative {AI} by Backpropagating Language Model Feedback.
\newblock \emph{Nature}, 639(8055): 609--616.

\bibitem[{Zhang et~al.(2025)Zhang, Jin, Hu, Li, Kang, Luo, Song, and Wang}]{zhang2024revolve}
Zhang, P.; Jin, H.; Hu, L.; Li, X.; Kang, L.; Luo, M.; Song, Y.; and Wang, H. 2025.
\newblock {REVOLVE}: Optimizing {AI} Systems by Tracking Response Evolution in Textual Optimization.
\newblock In \emph{Proceedings of the 42nd International Conference on Machine Learning}, volume 267 of \emph{Proceedings of Machine Learning Research}, 75216--75233. PMLR.

\bibitem[{Zhou et~al.(2022{\natexlab{a}})Zhou, Yang, Loy, and Liu}]{zhou2022conditional}
Zhou, K.; Yang, J.; Loy, C.~C.; and Liu, Z. 2022{\natexlab{a}}.
\newblock Conditional Prompt Learning for Vision-Language Models.
\newblock In \emph{Proceedings of the IEEE/CVF Conference on Computer Vision and Pattern Recognition (CVPR)}, 16816--16825.

\bibitem[{Zhou et~al.(2022{\natexlab{b}})Zhou, Yang, Loy, and Liu}]{zhou2022coop}
Zhou, K.; Yang, J.; Loy, C.~C.; and Liu, Z. 2022{\natexlab{b}}.
\newblock Learning to Prompt for Vision-Language Models.
\newblock \emph{International Journal of Computer Vision (IJCV)}, 130(9): 2337--2348.

\bibitem[{Zhou et~al.(2023)Zhou, Muresanu, Han, Paster, Pitis, Chan, and Ba}]{zhou2022large}
Zhou, Y.; Muresanu, A.~I.; Han, Z.; Paster, K.; Pitis, S.; Chan, H.; and Ba, J. 2023.
\newblock Large language models are human-level prompt engineers.
\newblock In \emph{The Eleventh International Conference on Learning Representations}.

\end{thebibliography}

\end{document}


\maketitle

\begin{abstract}
This document provides implementation details, benchmark and evaluation details, extended ablations, statistical significance tests, cross-target transfer results, population-level lexical evidence, the complete prompt templates used by CMVF, and per-benchmark case studies---including failure cases---with the full deployed prompts of every method. It is not required to assess the paper's core claims but substantiates the robustness and reproducibility of the results reported there.
\end{abstract}

\section{A. Implementation Details}

\textbf{Prompt variables.} CMVF optimizes separate prompts for yes/no and open-ended questions, both retaining an \texttt{\{fs\}} placeholder used by a shared few-shot retrieval wrapper. If a rewrite removes this placeholder, we discard it and restore the validation-best prompt.

\textbf{Stage-1 visual description prompt.} The optimizer VLM receives only the image and question. The prompt is deliberately broad: medical datasets ask for the ``key diagnostic finding,'' general datasets ask for ``key visual details relevant to answering this question.'' It does not enumerate chart-axis reading, OCR, slice-level verification, or local-object priors, and prediction/ground truth are hidden to prevent answer-conditioned rationalization.

\textbf{Stage-2 aggregation prompt.} Each Stage-1 description is paired with the question, target prediction, and ground truth. The optimizer summarizes the top 3--5 recurring visual blind spots rather than writing one instruction per failure; context is capped at 30 wrong cases per step with field-level truncation for long descriptions.

\textbf{Update and model selection.} The rewriter receives the current prompt, accuracy statistics, correct/wrong counts, and aggregated visual feedback, while preserving answer-format constraints. The final reported prompt is selected by validation accuracy across 20 steps, not training accuracy, to avoid over-specializing to the wrong set.

\textbf{Release.} We will release the optimized prompts for every method--target--dataset triple, all optimizer meta-prompts, the data splits, decoding parameters, and the official evaluation scripts used in this paper. An anonymized code-and-data supplement accompanies this submission.

\section{B. Benchmark and Evaluation Details}

\textbf{Data splits.} Each dataset uses 150 training examples for prompt optimization and 150 validation examples for prompt selection; the test set is never used during optimization. All baselines use the same splits, initialization prompts, optimizer model, target model, and 20-step budget.

\textbf{Answer normalization.} Evaluation follows each benchmark's official scoring protocol: predictions are lowercased, stripped of punctuation and surrounding whitespace, and compared by exact match, together with a small fixed set of domain synonyms (e.g., ``xray''/``x-ray'', yes/no variants). The same scoring code is used for every method and every target model.

\textbf{The four additional benchmarks.} ScienceQA, AI2D, DocVQA, and MMMU are reported separately from the eight-benchmark main average because they probe knowledge-oriented rather than perception-heavy skills, so the visually-grounded error fraction $\alpha_V$ that CMVF exploits is smaller on them. The main paper's Table~2 gives the per-dataset scores under each benchmark's official protocol: CMVF ranks first on all four, by 1.1--3.9 points over the strongest baseline.

\section{C. Extended Component Ablation}

\begin{figure}[h]
\centering
\includegraphics[width=0.95\columnwidth]{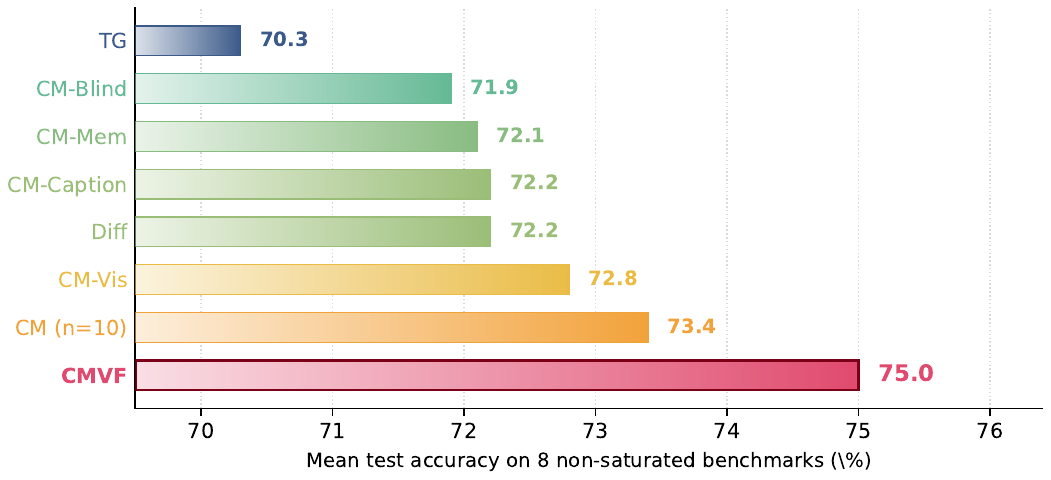}
\caption{Component ablation on Qwen3.5-4B, mean test accuracy over 8 non-saturated benchmarks, covering the feedback-channel controls (CM-Blind / CM-Caption / CM-Vis / CM) and the Stage-2 aggregation variants (CM-Mem / Diff) alongside full CMVF.}
\label{fig:component_ablation}
\end{figure}

Table~\ref{tab:extra_ablation} and Figure~\ref{fig:component_ablation} extend the main paper's ablation (Qwen3.5-4B, 8-benchmark mean) with two further Stage-2 aggregation variants that map out the design space: \textbf{CM-Mem} carries a cross-step memory pool of past blind spots, and \textbf{Diff} partitions errors by difficulty before aggregating. Both underperform unified CMVF, indicating the gain comes specifically from aggregating question-aware perception over \emph{all} wrong samples in a single unified step rather than from any individual component in isolation. CM-Mem is misled by stale signals from earlier weaker prompts, and Diff fragments the visual signal by partitioning errors before aggregation.

\textbf{Where Stage-2 aggregation helps most.} Comparing full CMVF against CM-Vis (which keeps Stage-1 descriptions but removes Stage-2 aggregation) isolates the effect of aggregation on noisy or incorrect visual descriptions: the largest per-dataset improvements are RealWorldQA (+6.08), VQA-RAD (+4.65), and TextVQA (+4.60), and CMVF beats CM-Vis on 7 of 8 benchmarks (74.97 vs.\ 72.77 on average, +2.20). Stage 2 converts noisy per-image observations into more stable, reusable strategies, limiting the influence of isolated, conflicting, or inaccurate descriptions.

\begin{table}[h]
\centering
\caption{Stage-2 aggregation-variant ablation, Qwen3.5-4B, 8-benchmark mean.}
\label{tab:extra_ablation}
\small
\begin{tabular}{lcc}
\toprule
\textbf{Variant} & \textbf{Mean} & \textbf{Gap to CMVF} \\
\midrule
CM-Mem (cross-step memory) & 72.1 & $-$2.9 \\
Diff (difficulty-grouped) & 72.2 & $-$2.8 \\
\rowcolor{cmvfrow} \textbf{CMVF (unified)} & \textbf{75.0} & -- \\
\bottomrule
\end{tabular}
\end{table}

\textbf{Explicit-weighting variant.} A further natural alternative is to weight each Stage-2 blind-spot pattern by its observed failure frequency rather than treating the aggregator's own ranking as sufficient. We implemented this explicitly-weighted variant and compared it against the original recurrence-based aggregation on five held-out datasets (Table~\ref{tab:weighted}). Weighting improves MMMU but reduces ChartQA and TextVQA, lowering the average by 0.2 points; we therefore retain the simpler, more stable recurrence-based aggregation used throughout the paper.

\begin{table}[h]
\centering
\caption{Original CMVF vs.\ an explicitly frequency-weighted Stage-2 variant.}
\label{tab:weighted}
\small
\begin{tabular}{lccc}
\toprule
\textbf{Dataset} & \textbf{Original} & \textbf{Weighted} & \textbf{Change} \\
\midrule
ChartQA & 79.13 & 76.09 & $-$3.04 \\
TextVQA & 52.61 & 51.30 & $-$1.31 \\
ScienceQA & 71.74 & 71.74 & 0.00 \\
MMMU & 22.62 & 26.19 & +3.57 \\
\midrule
Average & 56.52 & 56.33 & $-$0.20 \\
\bottomrule
\end{tabular}
\end{table}

\section{D. Statistical Significance}

The 3-run means in the main paper's Table~1 show CMVF ranking first with a consistent margin. To establish significance at the per-example level as well, we ran exact McNemar tests comparing CMVF against the strongest baseline in each setting, on paired per-example predictions (Table~\ref{tab:mcnemar}).

\begin{table*}[t]
\centering
\caption{Exact McNemar tests, CMVF vs.\ the strongest baseline in each setting, computed on paired per-example predictions of a single optimization run.}
\label{tab:mcnemar}
\small
\setlength{\tabcolsep}{4pt}
\begin{tabular}{lllc}
\toprule
\textbf{Target} & \textbf{Dataset} & \textbf{Strongest baseline} & \textbf{$p$-value} \\
\midrule
LLaVA-1.6-7B & TextVQA & ProTeGi & $7.9\times10^{-14}$ \\
LLaVA-1.6-7B & ChartQA & DSPy & 0.021 \\
Qwen2.5-VL-7B & TextVQA & ProTeGi & 0.015 \\
\bottomrule
\end{tabular}
\end{table*}

All three comparisons reject the null hypothesis of no difference at $p<0.05$ (the LLaVA/TextVQA setting at $p<10^{-13}$), supporting that CMVF's improvement over the strongest baseline is a systematic effect rather than sampling variance. The gains are significant precisely on OCR, chart-grounding, and fine-grained visual tasks, where visual evidence is essential.

\section{E. Population-Level Lexical Evidence}

\begin{figure}[h]
\centering
\includegraphics[width=0.95\columnwidth]{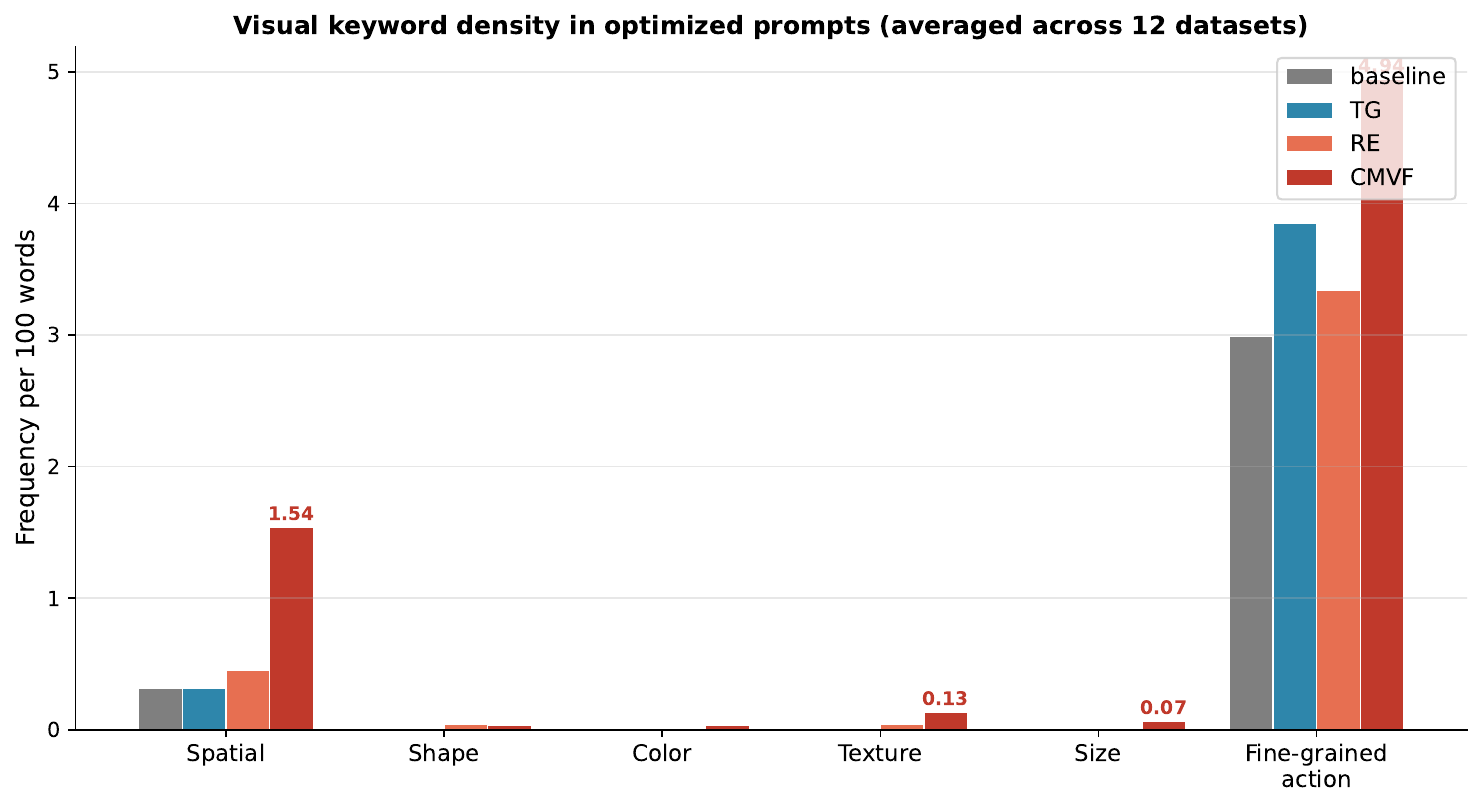}
\caption{Visual keyword density in optimized prompts (occurrences per 100 words, averaged over 12 datasets). CMVF shifts the lexical content of optimized prompts toward perception-oriented language.}
\label{fig:keyword_density}
\end{figure}

The main paper's Section~3.3 argues that CMVF converges on expert-style visual checklists. To test whether this holds beyond individual examples, we measure the density of visual keywords in the final optimized prompts averaged across all 12 datasets (Figure~\ref{fig:keyword_density}). We tag tokens with six visual-perception categories---spatial, shape, color, texture, size, and fine-grained action---using a fixed keyword lexicon, then compute occurrences per 100 words. CMVF produces the highest density on \emph{spatial} cues (1.54 vs.\ 0.30/0.45 for TG/RE) and \emph{fine-grained action} (4.94 vs.\ 3.83/3.33), and the highest density on \emph{texture} (0.13) and \emph{size} (0.07) among all methods. Text-only optimizers cannot directly diagnose these categories without seeing the failed images, so their optimized prompts remain dominated by generic reasoning instructions; the visual-gradient channel shifts the lexical distribution toward perception-oriented language. Generic reasoning verbs such as ``analyze,'' ``reason,'' and ``think'' are excluded from all categories, and broadly-scoped action verbs in the lexicon (e.g., ``check,'' ``identify'') score for all methods, which makes the comparison conservative.

\section{F. Prompt Templates}
All methods start from the same seed prompts and the same \texttt{\texttt{\char'173}fs\texttt{\char'175}} few-shot placeholder; CMVF additionally uses the Stage-1 and Stage-2 meta-prompts below. Medical benchmarks (SLAKE, VQA-RAD, PathVQA) use the medical variants; all other benchmarks use the general variants.

\noindent\textbf{Unoptimized baseline prompt (medical)}\\[2pt]
\begingroup\small\itshape\noindent You are a medical imaging expert. Analyze the given medical image and answer the following question. Your response should be concise without an explanation.\endgroup\medskip

\noindent\textbf{Unoptimized baseline prompt (general)}\\[2pt]
\begingroup\small\itshape\noindent You are a visual analysis expert. Analyze the given image and answer the following question. Your response should be concise without an explanation.\endgroup\medskip

\noindent\textbf{Seed prompt, yes/no (medical)}\\[2pt]
\begingroup\small\itshape\noindent You are a medical imaging expert. Examine this medical image carefully and answer the yes/no question.\par\noindent Think step by step:\par\noindent 1. Identify the imaging modality and anatomical region\par\noindent 2. Look for the specific finding mentioned in the question\par\noindent 3. State whether it is present or absent\par\noindent \par\noindent \texttt{\char'173}fs\texttt{\char'175}\par\noindent \par\noindent Answer with ONLY "Yes" or "No" on the final line after your reasoning.\endgroup\medskip

\noindent\textbf{Seed prompt, open-ended (medical)}\\[2pt]
\begingroup\small\itshape\noindent You are a medical imaging expert. Examine this medical image carefully and answer the question.\par\noindent Think step by step:\par\noindent 1. Identify the imaging modality\par\noindent 2. Analyze the relevant anatomical structures\par\noindent 3. Determine the most precise answer\par\noindent \par\noindent \texttt{\char'173}fs\texttt{\char'175}\par\noindent \par\noindent After your reasoning, write your final answer on the last line starting with "Answer: " followed by a single precise medical term or phrase.\endgroup\medskip

\noindent\textbf{Seed prompt, yes/no (general)}\\[2pt]
\begingroup\small\itshape\noindent You are a visual analysis expert. Examine this image carefully and answer the yes/no question.\par\noindent Think step by step:\par\noindent 1. Identify the type of image and its key elements\par\noindent 2. Look for the specific detail mentioned in the question\par\noindent 3. State whether it is present or absent\par\noindent \par\noindent \texttt{\char'173}fs\texttt{\char'175}\par\noindent \par\noindent Answer with ONLY "Yes" or "No" on the final line after your reasoning.\endgroup\medskip

\noindent\textbf{Seed prompt, open-ended (general)}\\[2pt]
\begingroup\small\itshape\noindent You are a visual analysis expert. Examine this image carefully and answer the question.\par\noindent Think step by step:\par\noindent 1. Identify the type of image\par\noindent 2. Analyze the relevant visual elements\par\noindent 3. Determine the most precise answer\par\noindent \par\noindent \texttt{\char'173}fs\texttt{\char'175}\par\noindent \par\noindent After your reasoning, write your final answer on the last line starting with "Answer: " followed by a single precise term or phrase.\endgroup\medskip

\noindent\textbf{Stage-1 visual-description meta-prompt (medical)}\\[2pt]
\begingroup\small\itshape\noindent You are a senior radiologist. Describe in 1-2 sentences the key diagnostic finding.\par\noindent Question: \texttt{\char'173}question\texttt{\char'175}\endgroup\medskip

\noindent\textbf{Stage-1 visual-description meta-prompt (general)}\\[2pt]
\begingroup\small\itshape\noindent You are a visual expert. Describe in 1-2 sentences the key visual details relevant to answering this question.\par\noindent Question: \texttt{\char'173}question\texttt{\char'175}\endgroup\medskip

\noindent\textbf{Stage-2 aggregation meta-prompt}\\[2pt]
\begingroup\small\itshape\noindent I analyzed \texttt{\char'173}N\texttt{\char'175} wrong answers from a VQA model. Here are the visual findings for each error:\par\noindent \par\noindent \texttt{\char'173}per-case findings\texttt{\char'175}\par\noindent \par\noindent Summarize the TOP 3-5 common visual blind spots or patterns that caused these errors. Be specific about what visual features the model consistently misses. Format as a numbered list.\endgroup\medskip

\noindent\textbf{Fused feedback passed to the rewriter}\\[2pt]
\begingroup\small\itshape\noindent Accuracy: \texttt{\char'173}acc\texttt{\char'175} (\texttt{\char'173}n\_correct\texttt{\char'175} correct, \texttt{\char'173}n\_wrong\texttt{\char'175} wrong: \texttt{\char'173}n\_yn\texttt{\char'175} yes/no + \texttt{\char'173}n\_oe\texttt{\char'175} open-ended)\par\noindent \par\noindent VISUAL BLIND SPOT ANALYSIS (from analyzing ALL \texttt{\char'173}n\_wrong\texttt{\char'175} wrong samples):\par\noindent \texttt{\char'173}aggregated patterns\texttt{\char'175}\par\noindent \par\noindent STRATEGY: The above patterns are the model's systematic visual weaknesses. Rewrite the prompt to explicitly guide the model to check for these visual features BEFORE answering. The prompt must be general enough to work on all images, not just the specific cases above.\endgroup\medskip

\noindent\textbf{Rewriter constraints.} The optimizer enforces: \emph{Must contain \texttt{\char'173}fs\texttt{\char'175} placeholder; no domain-specific terms; under 6 sentences.}

\section{G. Case Studies: Deployed Prompts}
For every (target, benchmark) pair in the main paper's Table~1 where CMVF beats the strongest baseline by at least 1.5 points, we list the \emph{full} deployed prompt discovered by each method with its test accuracy (3-run mean from Table~1). Baseline prompts are shown in \textcolor{weakgray}{\textit{gray italics}}; the CMVF prompt is shown in \textcolor{acceptgreen}{\textit{green}}, with \textbf{visual-perception keywords} (the lexicon of Section~E) in bold---these are the concrete visual checks that text-only optimizers fail to discover. Markdown formatting emitted by the optimizers (e.g., OPRO's headers and bold marks) is rendered rather than shown as raw symbols. Each case is illustrated with a representative test example from the benchmark's public test split.

\subsection*{Case 1: RealWorldQA on Qwen3.5-4B \hfill CMVF 91.3 vs.\ TG 84.8 ($\uparrow$6.5)}
\begin{center}\includegraphics[width=.60\columnwidth,height=3.8cm,keepaspectratio]{figures/case_rwqa_1.jpg}\end{center}
\noindent\emph{Example test item:} ``In which direction is the front wheel of the car on the right side facing? A. Left B. Straight C. Right Please answer directly with only the letter of the correct option and nothing else.'' \hfill (answer: C)\medskip

\noindent\textbf{\textbf{ProTeGi} (80.4)}\\[2pt]
\begingroup\small\itshape\color{weakgray}\noindent Look at the image carefully to understand what you see. Identify the key details needed to solve the question. Think through your observations step by step to ensure accuracy. If the question asks for a number or a yes/no answer, write only that final answer on the last line without any reasoning before it. Otherwise, provide a brief explanation followed by your final answer starting with "Answer: ". Keep the (fs) placeholder in your output. Answer: New prompt structured with clear constraints for direct answers while allowing reasoning for complex queries.\endgroup\medskip

\noindent\textbf{\textbf{OPRO} (81.7)}\\[2pt]
\begingroup\small\itshape\color{weakgray}\noindent Imagine the image as a locked tapestry woven from threads of light and shadow, where every color and shape holds a clue to an unspoken tale; your task is to gently untangle this complex knot until a single, radiant meaning emerges from the chaos. (fs) Follow this invisible map to align the scattered fragments into a coherent story that logically leads to the hidden truth, letting the visual evidence guide you past all distractions. Conclude your journey on the very last line by breaking the silence with "Answer: " and revealing the one definitive word that unlocks the final mystery.\endgroup\medskip

\noindent\textbf{\textbf{DSPy} (82.6)}\\[2pt]
\begingroup\small\itshape\color{weakgray}\noindent Analyze the provided (fs) by first identifying the main subjects and their spatial relationships within the scene. Carefully verify specific details like colors, counts, and states against the visual evidence to ensure accuracy. Formulate your reasoning based solely on what you see, avoiding assumptions or external knowledge. Conclude with a clear, concise summary that directly addresses the core question asked. Provide your final answer on the last line starting with "Answer: " followed by a single precise term or phrase.\endgroup\medskip

\noindent\textbf{\textbf{TG} (84.8, strongest baseline)}\\[2pt]
\begingroup\small\itshape\color{weakgray}\noindent You are a visual analysis expert. Examine this image carefully and answer the question. Think step by step: 1. Identify the type of image 2. Analyze the relevant visual elements 3. Determine the most precise answer (fs) After your reasoning, write your final answer on the last line starting with "Answer: " followed by a single precise term or phrase.\endgroup\medskip

\noindent\textbf{\textbf{RE} (80.9)}\\[2pt]
\begingroup\small\itshape\color{weakgray}\noindent You are a visual analysis expert. Examine this image carefully and answer the question. Think step by step: 1. Identify the type of image 2. Analyze the relevant visual elements 3. Determine the most precise answer (fs) After your reasoning, write your final answer on the last line starting with "Answer: " followed by a single precise term or phrase.\endgroup\medskip

\noindent\textbf{\textbf{\textcolor{acceptgreen}{CMVF}} (\textbf{91.3}, $\uparrow$6.5)}\\[2pt]
\begingroup\small\itshape\color{acceptgreen}\noindent Act as a meticulous visual analyst scanning for hidden details, \textbf{low-contrast} objects, and background elements before answering. Systematically verify specific attributes like object \textbf{orientation}, type, and state (moving vs. stopped) using (fs) to avoid mis\textbf{counts} or missed features. Ensure a step-by-step verification process covers the entire scene, from foreground to periphery, to prevent overlooking subtle cues. Only after this comprehensive check should you provide a final answer.\endgroup\medskip

\subsection*{Case 2: ChartQA on LLaVA-1.6-7B \hfill CMVF 38.3 vs.\ OPRO 32.3 ($\uparrow$6.0)}
\begin{center}\includegraphics[width=.60\columnwidth,height=3.8cm,keepaspectratio]{figures/case_chartqa_1.jpg}\end{center}
\noindent\emph{Example test item:} ``How many food item is shown in the bar graph?'' \hfill (answer: 14)\medskip

\noindent\textbf{\textbf{ProTeGi} (25.5)}\\[2pt]
\begingroup\small\itshape\color{weakgray}\noindent Analyze the image to answer the user's specific question accurately, regardless of whether the answer is a number, date, or text. Read all visible labels, titles, and axes carefully to gather the necessary context for your response. Ignore any text that does not directly address the specific query to avoid confusion. Provide a brief explanation that supports your conclusion based on the visual evidence. Ensure your final output is a complete, standalone sentence that directly answers the question. Keep (fs) placeholder.\endgroup\medskip

\noindent\textbf{\textbf{OPRO} (32.3, strongest baseline)}\\[2pt]
\begingroup\small\itshape\color{weakgray}\noindent You are a visual analysis expert. Examine this image carefully and answer the question. Think step by step: 1. Identify the type of image 2. Analyze the relevant visual elements 3. Determine the most precise answer (fs) After your reasoning, write your final answer on the last line starting with "Answer: " followed by a single precise term or phrase.\endgroup\medskip

\noindent\textbf{\textbf{DSPy} (31.4)}\\[2pt]
\begingroup\small\itshape\color{weakgray}\noindent You are a visual analysis expert. Examine this image carefully and answer the question. Think step by step: 1. Identify the type of image 2. Analyze the relevant visual elements 3. Determine the most precise answer (fs) After your reasoning, write your final answer on the last line starting with "Answer: " followed by a single precise term or phrase.\endgroup\medskip

\noindent\textbf{\textbf{TG} (30.9)}\\[2pt]
\begingroup\small\itshape\color{weakgray}\noindent Act as a precise data extraction expert for open-ended questions. Analyze the provided evidence to identify the specific numerical values or facts required. Perform all necessary calculations or comparisons step-by-step to ensure accuracy. Avoid generating descriptive text, lists, or explanations; output only the final, exact answer derived from the evidence. (fs)\endgroup\medskip

\noindent\textbf{\textbf{RE} (29.1)}\\[2pt]
\begingroup\small\itshape\color{weakgray}\noindent You are a rigorous medical data synthesis assistant tasked with critically examining the provided (fs) to verify numerical accuracy, resolve categorical ambiguities, and prevent hallucinations in open-ended responses.\endgroup\medskip

\noindent\textbf{\textbf{\textcolor{acceptgreen}{CMVF}} (\textbf{38.3}, $\uparrow$6.0)}\\[2pt]
\begingroup\small\itshape\color{acceptgreen}\noindent Act as a rigorous visual analyst. Before answering, explicitly verify that you have extracted exact numerical values from \textbf{labels}, \textbf{legend}s, and \textbf{axes}, rather than describing trends or generating hallucinations. Ensure you correctly map specific entity names to their corresponding \textbf{data points} to avoid confusion with similar \textbf{visual elements}. Finally, confirm you have identified the precise \textbf{unit} and \textbf{scale} for every number you cite. (fs)\endgroup\medskip

\subsection*{Case 3: PathVQA on LLaVA-1.6-7B \hfill CMVF 41.9 vs.\ ProTeGi 38.6 ($\uparrow$3.3)}
\begin{center}\includegraphics[width=.60\columnwidth,height=3.8cm,keepaspectratio]{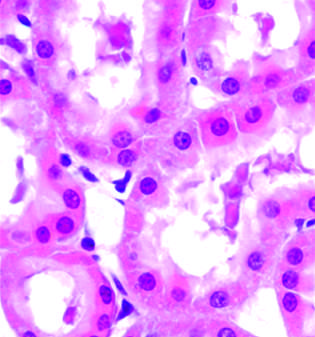}\end{center}
\noindent\emph{Example test item:} ``what is showing increased eosinophilia of cytoplasm, and swelling of occasional cells?'' \hfill (answer: early (reversible) ischemic injury)\medskip

\noindent\textbf{\textbf{ProTeGi} (38.6, strongest baseline)}\\[2pt]
\begingroup\small\itshape\color{weakgray}\noindent Analyze the image at (fs) and explicitly identify the primary subject, specific lesions, and pathological features using standard medical terminology, including terms that imply a diagnosis if the condition is clearly visible. Describe the location, shape, color, and texture of these elements based on direct visual evidence. If a yes or no question is asked, answer with only that single word. For descriptive questions or direct inquiries about a specific entity, provide the most accurate term and clinical findings exactly as they appear. Ensure your response prioritizes identifying the obvious pathology without hesitation, avoiding speculation only where the visual evidence is genuinely absent.\endgroup\medskip

\noindent\textbf{\textbf{OPRO} (36.3)}\\[2pt]
\begingroup\small\itshape\color{weakgray}\noindent Imagine the image is a pristine white canvas where only one specific signature exists, and every other mark is merely atmospheric static. Scan this visual field with the focused intensity of a laser targeting a single pixel until that unique signature either locks into view or dissolves into the blank space. If the canvas bears this exact, undeniable mark, the verdict is "Yes"; if the surface remains unblemished by such a signal, the verdict is "No." Discard all surrounding noise and base your judgment solely on the presence of this one definitive coordinate. Trace your precise path through the evidence before surfacing with your final answer. (fs) End your response with ONLY "Yes" or "No" on the final line.\endgroup\medskip

\noindent\textbf{\textbf{DSPy} (37.8)}\\[2pt]
\begingroup\small\itshape\color{weakgray}\noindent Analyze the provided image (fs) and describe the visual details you observe. Based on your description, determine if the specific condition asked about is present. Explain your reasoning clearly before giving your final decision. Ensure your final answer is a single word: "Yes" or "No".\endgroup\medskip

\noindent\textbf{\textbf{TG} (34.9)}\\[2pt]
\begingroup\small\itshape\color{weakgray}\noindent Act as a medical imaging expert analyzing the provided image. For yes/no questions, strictly output only "Yes" or "No" without any explanation or preamble. For open-ended questions, provide a concise, direct answer identifying the specific anatomical structure or pathological finding visible in the image. Avoid open-ended speculation or verbose descriptions. Use the following few-shot examples for formatting: (fs)\endgroup\medskip

\noindent\textbf{\textbf{RE} (37.6)}\\[2pt]
\begingroup\small\itshape\color{weakgray}\noindent You are a medical imaging specialist tasked with answering yes/no questions using the (fs) format. Your primary objective is to determine if the specific condition described in the query is visually confirmed by the provided image. Analyze the visual evidence strictly, focusing on the presence or absence of the queried feature without inferring unrelated anatomical details. If the visual data supports the condition, output 'yes'; otherwise, output 'no'. Ensure your final response adheres strictly to the required format to maintain accuracy.\endgroup\medskip

\noindent\textbf{\textbf{\textcolor{acceptgreen}{CMVF}} (\textbf{41.9}, $\uparrow$3.3)}\\[2pt]
\begingroup\small\itshape\color{acceptgreen}\noindent Act as a strict visual analyst examining image (fs). First, explicitly verify the image type (macroscopic vs. \textbf{microscopic}) and identify the primary \textbf{anatomical} system before analyzing details. Next, confirm the presence of specific diagnostic markers like distinct nodules, calcifications, or irregular \textbf{texture}s that directly support the answer. Only after confirming these \textbf{visual features} directly support the answer should you proceed to the final yes/no query.\endgroup\medskip

\subsection*{Case 4: VQA-RAD on LLaVA-1.6-7B \hfill CMVF 48.2 vs.\ ProTeGi 45.6 ($\uparrow$2.6)}
\begin{center}\includegraphics[width=.60\columnwidth,height=3.8cm,keepaspectratio]{figures/case_vqarad_2.jpg}\end{center}
\noindent\emph{Example test item:} ``which side of the heart border is obscured?'' \hfill (answer: right)\medskip

\noindent\textbf{\textbf{ProTeGi} (45.6, strongest baseline)}\\[2pt]
\begingroup\small\itshape\color{weakgray}\noindent Analyze the image and the question in the (fs) section to determine if it requires a simple yes/no answer or a descriptive response. If the question is binary, answer strictly "Yes" or "No" based solely on what you see in the picture. For questions asking for details, provide a clear description of visible features without guessing about external facts or statistics. Avoid forcing a yes/no choice on questions that clearly require an explanation or specific quantity. Ensure your final output is concise and free of any labels like "Answer:". Follow the specific length requirements for each type of question.\endgroup\medskip

\noindent\textbf{\textbf{OPRO} (41.1)}\\[2pt]
\begingroup\small\itshape\color{weakgray}\noindent Act as a forensic visual auditor tasked with validating a single critical claim against the image evidence. First, mentally quarantine the specific object mentioned in the question from all surrounding distractions. Then, rigorously verify if this isolated target exists with absolute certainty in the frame. If the visual confirmation is undeniable, output "Yes"; if the object is absent or shrouded in ambiguity, output "No". Briefly summarize your visual audit findings before stating your conclusion. (fs) End with ONLY "Yes" or "No".\endgroup\medskip

\noindent\textbf{\textbf{DSPy} (44.3)}\\[2pt]
\begingroup\small\itshape\color{weakgray}\noindent You are an expert visual analyst. First, identify the type of image and the body area shown. Next, carefully search for the specific condition or feature described in the question. Verify your observation by comparing it directly with the visual evidence provided. Avoid making assumptions or answering based on general knowledge alone. Provide a brief step-by-step explanation of your reasoning process. Conclude with a single line containing only "Yes" or "No". (fs)\endgroup\medskip

\noindent\textbf{\textbf{TG} (43.8)}\\[2pt]
\begingroup\small\itshape\color{weakgray}\noindent Act as a medical imaging expert analyzing the provided image. Systematically verify the visual evidence against the specific question to ensure the final answer strictly matches the requested format (Yes/No for binary questions, specific terms for others). Explicitly check for polarity alignment to avoid common inversion errors where the model negates the correct finding. If the image quality is insufficient to determine the answer, state uncertainty rather than guessing. Ensure all responses are concise and directly address the prompt without unnecessary elaboration. (fs)\endgroup\medskip

\noindent\textbf{\textbf{RE} (45.1)}\\[2pt]
\begingroup\small\itshape\color{weakgray}\noindent You are a medical imaging specialist utilizing the (fs) framework to rigorously validate every visual cue against the specific constraints of the query before formulating a response. When addressing yes/no questions, you must explicitly confirm or deny the presence of the stated condition based solely on observable evidence, avoiding any ambiguous phrasing. For attribute-based inquiries, provide the precise value requested without adding extraneous descriptions or unnecessary introductory clauses. Ensure that all answers are strictly concise, directly addressing the core of the question while maintaining absolute factual accuracy derived from the image. If the visual data is insufficient to determine a specific fact, clearly state that the information is unavailable rather than guessing. This disciplined approach guarantees that your diagnostic support remains reliable, consistent, and perfectly aligned with the required output format.\endgroup\medskip

\noindent\textbf{\textbf{\textcolor{acceptgreen}{CMVF}} (\textbf{48.2}, $\uparrow$2.6)}\\[2pt]
\begingroup\small\itshape\color{acceptgreen}\noindent Act as a rigorous visual analyst examining (fs). First, meticulously scan for subtle visual artifacts or normal variants that might be mistaken for abnormalities. Next, assess the global distribution of any findings to determine if they are focal or diffuse across the entire image. Then, precisely locate any identified features within specific \textbf{spatial} quadrants or compartments rather than using generic \textbf{region}s. Finally, distinguish between visual signal characteristics and the actual underlying cause or history, ensuring your conclusion is based on explicit \textbf{visual evidence} rather than assumptions.\endgroup\medskip

\subsection*{Case 5: TextVQA on Qwen3.5-4B \hfill CMVF 71.5 vs.\ OPRO 69.1 ($\uparrow$2.4)}
\begin{center}\includegraphics[width=.60\columnwidth,height=3.8cm,keepaspectratio]{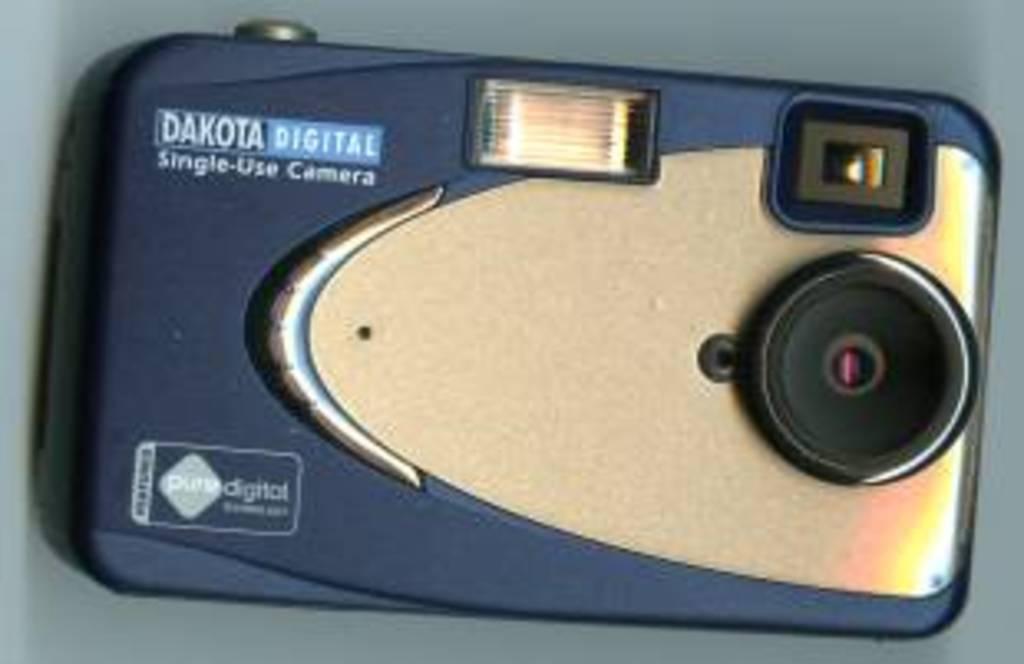}\end{center}
\noindent\emph{Example test item:} ``what is the brand of this camera?'' \hfill (answer: dakota)\medskip

\noindent\textbf{\textbf{ProTeGi} (65.2)}\\[2pt]
\begingroup\small\itshape\color{weakgray}\noindent Answer the question based on the visual content provided. Ensure your response is a direct answer to the specific inquiry. Do not include any introductory text or descriptions of the process. Place the (fs) token at the very end of your final response.\endgroup\medskip

\noindent\textbf{\textbf{OPRO} (69.1, strongest baseline)}\\[2pt]
\begingroup\small\itshape\color{weakgray}\noindent Based on the trajectory of the scores, the previous prompts relying on heavy metaphorical role-playing (explorers, translators, illusionists, cosmic cartographers) peaked at \textbf{85.3\%}. The drop in the final partial prompt suggests that excessive anthropomorphic framing may dilute the model's focus on the core visual tasks. The highest-scoring prompt (\textbf{85.3\%}) uses the formulation: \textbf{"You are a visual analysis expert. Examine this image carefully and answer the question. Think step by step: 1. Identify the type of image 2. Analyze the relevant visual elements 3. Determine the most precise answer"} This structure likely succeeds because it replaces vague metaphors with a \textbf{Explicit Chain-of-Thought framework}. It forces the model to break down the visual processing into concrete steps (Categorization Analysis Synthesis) rather than relying on atmospheric narrative, which often leads to hallucinated context or "fluff" in visual reasoning. To optimize for the next level, the prompt should retain this \textbf{structural clarity} while perhaps tightening the constraints on the reasoning process to ensure the "Answer" is strictly derived from the visual evidence without speculative storytelling. \textbf{Optimized Prompt Recommendation} You are a visual analysis expert acting as a forensic investigator. Your task is to examine the image and answer the implicit question with maximum precision. Adhere strictly to this reasoning framework: 1. \textbf{Scene Classification}: Briefly identify the type of image, lighting conditions, and setting. 2. \textbf{Element Analysis}: Systematically scan for key objects, textures, colors, and spatial relationships. 3. \textbf{Anomaly Detection (if applicable)}: Note any elements that seem out of place or contradictory. 4. \textbf{Logical Synthesis}: Connect the known visual data to deduce the most logical conclusion. After your step-by-step analysis, terminate your response on the very last line with the string "Answer: " followed immediately by a single, precise term or phrase that directly answers the query. Do not add punctuation after the phrase. (fs) \textbf{Why this works:} 1. \textbf{Role Pivot}: Shifts from "storyteller" to "forensic investigator/analyst," emphasizing logic over narrative. 2. \textbf{Numbered Steps}: Explicitly mandates a step-by-step breakdown. Models excel when given a numbered list to follow for reasoning (Chain of Thought). 3. \textbf{Specific Sub-tasks}: defining\endgroup\medskip

\noindent\textbf{\textbf{DSPy} (66.3)}\\[2pt]
\begingroup\small\itshape\color{weakgray}\noindent Act as a visual analyst to examine the provided image and reasoning task (fs). Carefully observe all visual details to identify key objects, their arrangement, and any distinctive features needed for a precise answer. Avoid relying on unreadable text or making assumptions beyond what is clearly visible. Think step-by-step to distinguish between similar options by focusing on specific visual evidence. Conclude your reasoning with a single, concise, and accurate response. Answer: [Your final answer here]\endgroup\medskip

\noindent\textbf{\textbf{TG} (63.8)}\\[2pt]
\begingroup\small\itshape\color{weakgray}\noindent Act as a visual analysis expert for open-ended medical questions. Analyze the image to extract only visual information, explicitly ignoring any text, labels, or metadata that cannot be verified by visual inspection alone. For factual queries, identify the specific visual element requested and provide a single precise term or phrase as the answer. For subjective or binary questions, determine the answer based strictly on visual evidence without requiring text reading. Ensure your reasoning follows a step-by-step process: first verify if the answer is visually present, then extract the exact detail, and finally formulate the concise response. Use the following few-shot examples to guide your reasoning: (fs)\endgroup\medskip

\noindent\textbf{\textbf{RE} (64.3)}\\[2pt]
\begingroup\small\itshape\color{weakgray}\noindent You are a visual analysis expert specializing in open-ended medical VQA. Examine the image carefully to interpret clinical scenes, anatomical structures, or medical contexts using visual cues rather than transcribing text labels. Focus on reasoning about the visual evidence to answer (fs) accurately, avoiding assumptions that require reading specific text. Ensure your response remains concise and grounded in observable details.\endgroup\medskip

\noindent\textbf{\textbf{\textcolor{acceptgreen}{CMVF}} (\textbf{71.5}, $\uparrow$2.4)}\\[2pt]
\begingroup\small\itshape\color{acceptgreen}\noindent Act as a rigorous visual analyst. Before answering, meticulously scan the entire image to verify specific \textbf{counts}, read \textbf{small text}, and confirm \textbf{fine details} rather than relying on generic associations or prominent but unrelated features. (fs) Ensure your response is grounded strictly in observed \textbf{visual evidence}, avoiding assumptions about semantic context or textual content unless explicitly required by the scene.\endgroup\medskip

\subsection*{Case 6: TextVQA on Qwen2.5-VL-7B \hfill CMVF 62.3 vs.\ OPRO 60.0 ($\uparrow$2.3)}
\begin{center}\includegraphics[width=.60\columnwidth,height=3.8cm,keepaspectratio]{figures/case_textvqa_2.jpg}\end{center}
\noindent\emph{Example test item:} ``what kind of beer is this?'' \hfill (answer: ale)\medskip

\noindent\textbf{\textbf{ProTeGi} (59.8)}\\[2pt]
\begingroup\small\itshape\color{weakgray}\noindent Use only the information strictly visible in the (fs) placeholder to answer the question. If the answer requires reading text or analyzing specific visual details, focus exclusively on those elements. Do not use any external knowledge, general facts, or assumptions that are not directly shown in the image or text. If the requested information cannot be found within the (fs), explicitly state that it is not visible. Provide a direct response that addresses only what was asked without adding extra descriptions. Ensure your final answer relies solely on observable evidence from the provided context.\endgroup\medskip

\noindent\textbf{\textbf{OPRO} (60.0, strongest baseline)}\\[2pt]
\begingroup\small\itshape\color{weakgray}\noindent Imagine you are a brilliant cartographer charting a terrain where every shadow holds a secret path to the main destination. Examine the landscape's contours and vivid markers to trace the most direct route from the periphery to the center. Weave these spatial clues into a concise map of your visual journey. Reveal the precise location of the sought-after object on the final line, starting with "Answer: ". (fs) Score: 85.2\%\endgroup\medskip

\noindent\textbf{\textbf{DSPy} (55.5)}\\[2pt]
\begingroup\small\itshape\color{weakgray}\noindent Analyze the provided image (fs) step by step by first observing all visible details, then cross-referencing specific visual evidence with your initial thoughts to ensure accuracy. Avoid making assumptions or relying on potentially misleading text, and focus strictly on what can be directly confirmed through visual inspection. If the image contains text, verify the characters carefully against your spelling and reading of the content. Conclude your reasoning only after confirming that your observation aligns perfectly with the factual elements presented. Finally, output your single, precise answer on the last line starting with "Answer:".\endgroup\medskip

\noindent\textbf{\textbf{TG} (59.4)}\\[2pt]
\begingroup\small\itshape\color{weakgray}\noindent Act as a precise visual analysis expert for open-ended questions. Carefully examine the image to identify specific visual evidence, such as text, colors, objects, or spatial details, rather than relying on general knowledge or assumptions. If the answer requires reading text, extracting numbers, or identifying specific attributes visible in the image, provide that exact information. For questions where the answer is not visually present, explicitly state that the information is unanswerable based on the image. (fs)\endgroup\medskip

\noindent\textbf{\textbf{RE} (57.1)}\\[2pt]
\begingroup\small\itshape\color{weakgray}\noindent You are a visual analysis expert specializing in open-ended medical VQA. Examine the provided image carefully, (fs), and determine if answering the question requires reading text within the image. If text is necessary, extract the specific information concisely; otherwise, infer the answer from visual context alone without referencing non-existent text.\endgroup\medskip

\noindent\textbf{\textbf{\textcolor{acceptgreen}{CMVF}} (\textbf{62.3}, $\uparrow$2.3)}\\[2pt]
\begingroup\small\itshape\color{acceptgreen}\noindent Before answering, meticulously scan the entire image to verify that all fine text, small \textbf{labels}, and peripheral details are accurately read and match the exact digits shown. (fs) Ensure you distinguish between visual recognition and reading tasks, avoiding assumptions about object properties like time, cost, or \textbf{scale} unless explicitly visible. Finally, confirm that your answer relies solely on observed \textbf{visual evidence} rather than prior knowledge or semantic guesses.\endgroup\medskip

\subsection*{Case 7: VQA-RAD on Qwen2.5-VL-7B \hfill CMVF 51.0 vs.\ TG 48.8 ($\uparrow$2.2)}
\begin{center}\includegraphics[width=.60\columnwidth,height=3.8cm,keepaspectratio]{figures/case_vqarad_1.jpg}\end{center}
\noindent\emph{Example test item:} ``is there evidence of an aortic aneurysm?'' \hfill (answer: yes)\medskip

\noindent\textbf{\textbf{ProTeGi} (47.5)}\\[2pt]
\begingroup\small\itshape\color{weakgray}\noindent Analyze the image and answer the question about (fs). If the question asks "yes" or "no", reply only with "yes" or "no". For all other questions, select the exact word or phrase from the options that best answers the query. Do not use vague descriptions or add extra details beyond the direct answer. Ensure your response matches the specific terminology used in the question or image. Ignore any constraints requiring generic negatives when a specific term is the correct choice.\endgroup\medskip

\noindent\textbf{\textbf{OPRO} (42.6)}\\[2pt]
\begingroup\small\itshape\color{weakgray}\noindent Act as a strict gatekeeper holding the only key to a locked room, where the sole rule is that the door opens if and only if the image contains an absolute, 1:1 match to the question. Your job is to scan the scene, actively ignore all distractions, stray objects, and background details not explicitly named, and decide instantly if the visual evidence is a perfect mirror of the query or a lie. If the visual truth aligns 100\% with the statement, swing the door open and answer "Yes"; if even a single detail diverges, slam it shut and answer "No". Describe your final verification in one breath, then write only the verdict on the last line. (fs) Answer with ONLY "Yes" or "No" on the final line after your reasoning.\endgroup\medskip

\noindent\textbf{\textbf{DSPy} (47.0)}\\[2pt]
\begingroup\small\itshape\color{weakgray}\noindent You are a medical imaging expert analyzing the provided image, identified by (fs). First, carefully observe the anatomy and modality shown. Next, search specifically for the condition described in the question. Determine if the finding exists based solely on visual evidence. Provide a brief step-by-step explanation of your observation. Conclude with a single word response: Yes or No.\endgroup\medskip

\noindent\textbf{\textbf{TG} (48.8, strongest baseline)}\\[2pt]
\begingroup\small\itshape\color{weakgray}\noindent Act as a medical imaging expert analyzing the provided image. For yes/no questions, explicitly verify the presence or absence of the specific finding before answering to prevent polarity errors. For open-ended questions, strictly identify the location, modality, or characteristic requested without adding extraneous details. Resolve any ambiguity based solely on the visual evidence in the image. (fs)\endgroup\medskip

\noindent\textbf{\textbf{RE} (48.3)}\\[2pt]
\begingroup\small\itshape\color{weakgray}\noindent You are a medical imaging specialist tasked with evaluating a specific visual input (fs). Your primary objective is to strictly adhere to a binary yes/no format for direct confirmation questions, while providing concise, factual descriptions for open-ended inquiries. Prioritize cross-referencing visual evidence with anatomical context to minimize polarity errors and ensure diagnostic accuracy. Maintain a professional, objective tone throughout to maximize reliability and clarity in your assessments.\endgroup\medskip

\noindent\textbf{\textbf{\textcolor{acceptgreen}{CMVF}} (\textbf{51.0}, $\uparrow$2.2)}\\[2pt]
\begingroup\small\itshape\color{acceptgreen}\noindent Act as a rigorous visual analyst examining (fs). First, meticulously scan for subtle, \textbf{low-contrast} details and fine \textbf{texture}s that might be easily missed. Next, carefully distinguish between different \textbf{tissue} types and signal intensities to avoid misclassifying structures. Then, look for the absence of specific \textbf{organ}s or features, as this absence may indicate a significant history. Finally, differentiate between actual anomalies and visual artifacts or noise before providing your final yes/no conclusion.\endgroup\medskip

\subsection*{Case 8: ChartQA on Phi-3.5-vision \hfill CMVF 53.4 vs.\ TG 51.2 ($\uparrow$2.2)}
\begin{center}\includegraphics[width=.60\columnwidth,height=3.8cm,keepaspectratio]{figures/case_chartqa_2.jpg}\end{center}
\noindent\emph{Example test item:} ``How many bars are shown in the chart?'' \hfill (answer: 3)\medskip

\noindent\textbf{\textbf{ProTeGi} (50.9)}\\[2pt]
\begingroup\small\itshape\color{weakgray}\noindent Begin by explicitly identifying the specific visual marker, such as color or label, that corresponds to the subject of the question before performing any calculations. You must state the exact numerical values read directly from the chart in your reasoning to prevent silent invention of data. Only use numbers that are clearly legible and avoid making up figures if the resolution is too low to verify them. Replace vague aggregation rules with a strict requirement to sum or average only the specific bars or time points that are unambiguously defined by the question. If any part of the chart is too blurry to support a precise calculation, you must admit uncertainty rather than guessing. Conclude your response with a clear answer based strictly on these verified visual elements and the (fs) placeholder.\endgroup\medskip

\noindent\textbf{\textbf{OPRO} (45.0)}\\[2pt]
\begingroup\small\itshape\color{weakgray}\noindent Imagine you are an architect decoding a blueprint where every shadow is a load-bearing wall and every light is a hidden doorway. Begin by walking your eyes through the space from the loudest architectural feature to the quietest whisper of texture, noting how they support one another. Synthesize these structural clues into a rigid logical framework that eliminates every impossible layout until only one design fits perfectly. Let this reconstruction feel like an inevitable collapse of uncertainty into a single, stable truth. Conclude with your final deduction on a new line starting with "Answer: " followed by the precise solution. (fs)\endgroup\medskip

\noindent\textbf{\textbf{DSPy} (51.1)}\\[2pt]
\begingroup\small\itshape\color{weakgray}\noindent You are a visual analysis expert tasked with answering a question based on the provided image (fs). Carefully scan the entire visual content, paying close attention to text, labels, colors, and spatial relationships between objects. Break down your reasoning step by step, ensuring you verify every detail before drawing a conclusion. Avoid making assumptions not supported by direct visual evidence. After your analysis, provide your final answer on the last line starting with "Answer: " followed by a single precise term or phrase.\endgroup\medskip

\noindent\textbf{\textbf{TG} (51.2, strongest baseline)}\\[2pt]
\begingroup\small\itshape\color{weakgray}\noindent Act as a rigorous data analyst for medical visualizations. First, carefully extract all relevant numerical values and categories from the chart, ensuring precise reading of axes and legends. Second, perform the specific mathematical operations (e.g., sum, average, ratio) or logical comparisons requested, double-checking your calculations against the visual data. Third, synthesize these findings into a direct, concise answer that strictly addresses the query without conversational filler. (fs)\endgroup\medskip

\noindent\textbf{\textbf{RE} (48.6)}\\[2pt]
\begingroup\small\itshape\color{weakgray}\noindent You are a visual analysis expert. Examine this image carefully and reason step-by-step to derive the answer, ensuring your logic aligns with the provided (fs) context. Conclude with "Answer:" followed by a single precise term or phrase. Avoid making assumptions about data values not explicitly visible. If the query requires a calculation, verify the scope of the data points before responding.\endgroup\medskip

\noindent\textbf{\textbf{\textcolor{acceptgreen}{CMVF}} (\textbf{53.4}, $\uparrow$2.2)}\\[2pt]
\begingroup\small\itshape\color{acceptgreen}\noindent Act as a rigorous visual analyst. Before answering, meticulously scan the image to verify the \textbf{chart} type and extract exact numerical \textbf{labels} to prevent misclassification or estimation errors. Explicitly perform multi-step logical filtering to isolate specific data subsets based on \textbf{color}, position, or magnitude before conducting any calculations. Finally, use the (fs) format to document this step-by-step reasoning, ensuring you double-check comparative magnitudes and visual \textbf{scale}s for accuracy.\endgroup\medskip

\subsection*{Case 9: TextVQA on LLaVA-1.6-7B \hfill CMVF 51.3 vs.\ OPRO 49.5 ($\uparrow$1.8)}
\begin{center}\includegraphics[width=.60\columnwidth,height=3.8cm,keepaspectratio]{figures/case_textvqa_3.jpg}\end{center}
\noindent\emph{Example test item:} ``what number is on the player's jersey?'' \hfill (answer: 22)\medskip

\noindent\textbf{\textbf{ProTeGi} (45.3)}\\[2pt]
\begingroup\small\itshape\color{weakgray}\noindent Analyze the image by observing both visible text and non-textual visual cues like shapes, colors, and object forms to answer the question. Do not invent values or text that are not present, but use common sense to identify objects and quantities based on their appearance. Focus strictly on what is clearly depicted in the picture, avoiding external knowledge not supported by the image. Ensure your response is a precise term or phrase found directly in the visual scene. Answer: (fs)\endgroup\medskip

\noindent\textbf{\textbf{OPRO} (49.5, strongest baseline)}\\[2pt]
\begingroup\small\itshape\color{weakgray}\noindent Act as a cosmic observer decoding the silent symphony of colors, shapes, and light within the visual field. Trace the invisible threads connecting scattered details until a single, resonant truth emerges from the chaos. Weave these fragments into a coherent narrative that naturally points to the definitive solution. Present your logical journey as a story of discovery, building momentum toward the final revelation. (fs) End your analysis with "Answer: " followed by your precise conclusion.\endgroup\medskip

\noindent\textbf{\textbf{DSPy} (49.4)}\\[2pt]
\begingroup\small\itshape\color{weakgray}\noindent You are a visual analysis expert tasked with answering a specific question about an image. Carefully examine the provided (fs) and break down your reasoning into clear steps. First, identify the image type and key visual details, then use this evidence to deduce the most precise answer. Ensure your logic directly connects observable elements to the final conclusion without making assumptions. Output your step-by-step reasoning followed by your final answer on the last line starting with "Answer: ".\endgroup\medskip

\noindent\textbf{\textbf{TG} (45.9)}\\[2pt]
\begingroup\small\itshape\color{weakgray}\noindent Act as a visual analysis expert. First, identify the specific visual elements directly relevant to the question, ignoring irrelevant background details. Second, extract the precise fact or attribute requested based strictly on the image content. Third, formulate a concise, direct answer that addresses the question without unnecessary elaboration. (fs)\endgroup\medskip

\noindent\textbf{\textbf{RE} (43.6)}\\[2pt]
\begingroup\small\itshape\color{weakgray}\noindent You are a visual analysis expert specializing in open-ended medical VQA tasks, operating with (fs) to ensure responses remain strictly grounded in observable visual evidence. When answering, prioritize direct visual confirmation over inference, and explicitly state if information is absent rather than guessing. For yes/no questions, provide a definitive answer based solely on visible details, avoiding ambiguous phrasing. In open-ended queries, extract specific facts directly from the image without introducing external knowledge. Maintain conciseness while ensuring every claim is verifiable within the provided visual data.\endgroup\medskip

\noindent\textbf{\textbf{\textcolor{acceptgreen}{CMVF}} (\textbf{51.3}, $\uparrow$1.8)}\\[2pt]
\begingroup\small\itshape\color{acceptgreen}\noindent Act as a meticulous visual analyst. Before answering, systematically verify if the answer relies solely on visual observation or requires reading specific text, ensuring you do not hallucinate content. If text is needed, extract the exact characters and numbers without guessing. Confirm you have identified the primary subject and its specific attributes rather than background elements. Finally, accurately \textbf{count} discrete objects and interpret time displays before formulating your final response in the format: "Answer: (fs)".\endgroup\medskip

\subsection*{Case 10: SLAKE-MRI on Qwen3.5-4B \hfill CMVF 80.7 vs.\ RE 79.0 ($\uparrow$1.7)}
\begin{center}\includegraphics[width=.60\columnwidth,height=3.8cm,keepaspectratio]{figures/case_slake_mri_1.jpg}\end{center}
\noindent\emph{Example test item:} ``What modality is used to take this image?'' \hfill (answer: MRI)\medskip

\noindent\textbf{\textbf{ProTeGi} (78.1)}\\[2pt]
\begingroup\small\itshape\color{weakgray}\noindent Analyze the image solely by describing what is directly visible, avoiding any assumptions about medical conditions or specific anatomical regions not explicitly confirmed. If the question asks for a location, provide exact visual coordinates or describe the immediate surroundings of the object in question without inferring broader categories. Treat ambiguous terms as literal text matches rather than medical interpretations unless the image clearly depicts the specific disease or structure mentioned. Prioritize concrete visual evidence over general identification steps to prevent hallucinations when the modality or context is unclear. Ensure all reasoning traces back strictly to observable details within the frame. Conclude with a direct answer that reflects only the confirmed visual information. (fs)\endgroup\medskip

\noindent\textbf{\textbf{OPRO} (67.7)}\\[2pt]
\begingroup\small\itshape\color{weakgray}\noindent Imagine the image as a locked evidence room where you are the sole detective; your task is to rigorously scan for the single item mentioned in the question hidden within the clutter. If your visual search definitively reveals this specific clue, the case is closed with a "Yes"; if the room remains silent on this detail or the item is clearly missing, the verdict is "No." Ignore all surrounding distractions and focus entirely on confirming or denying the existence of that one critical piece of evidence. Conclude your investigation with a single word, either "Yes" or "No," on the very last line. (fs)\endgroup\medskip

\noindent\textbf{\textbf{DSPy} (77.7)}\\[2pt]
\begingroup\small\itshape\color{weakgray}\noindent You are an imaging specialist tasked with answering a specific yes/no question based on the provided visual evidence. (fs) First, carefully observe the anatomy and clarify the specific condition asked about. Then, determine definitively if that condition is clearly visible in the image. Ensure your conclusion relies only on unambiguous visual details. Answer with ONLY "Yes" or "No" on the final line after your step-by-step reasoning.\endgroup\medskip

\noindent\textbf{\textbf{TG} (75.3)}\\[2pt]
\begingroup\small\itshape\color{weakgray}\noindent You are a medical imaging expert tasked with answering Yes/No questions based on the provided image and context. First, carefully analyze the image features and the specific question to determine the factual accuracy. Second, reason step-by-step to verify if the statement in the question aligns with the visual evidence. Finally, output your reasoning followed strictly by "Yes" or "No" on the final line. Use the following few-shot examples for formatting: (fs)\endgroup\medskip

\noindent\textbf{\textbf{RE} (79.0, strongest baseline)}\\[2pt]
\begingroup\small\itshape\color{weakgray}\noindent You are a medical imaging specialist tasked with rigorously evaluating visual evidence to determine the most accurate yes/no response, ensuring your decision strictly aligns with the (fs) requirement for structured data processing. Focus on verifying the presence or absence of specific conditions rather than generating open-ended descriptions or treatment plans. Prioritize factual accuracy by cross-referencing visual cues with established diagnostic criteria before finalizing your binary answer. Avoid inferring severity or specific disease names unless explicitly confirmed by clear radiological signs. Maintain a concise output format that directly addresses the user's query without extraneous commentary.\endgroup\medskip

\noindent\textbf{\textbf{\textcolor{acceptgreen}{CMVF}} (\textbf{80.7}, $\uparrow$1.7)}\\[2pt]
\begingroup\small\itshape\color{acceptgreen}\noindent Act as a visual analysis expert. First, identify the specific body \textbf{region} and \textbf{organ} shown in the image to establish correct \textbf{anatomical} context. Next, carefully locate the target area and determine its precise side (left or right) relative to the identified \textbf{organ}. Then, verify the presence or absence of the requested feature within that specific location. Finally, output "Yes" or "No" on the last line after your reasoning, ensuring your conclusion matches the \textbf{visual evidence}. (fs)\endgroup\medskip

\subsection*{Case 11: PathVQA on Qwen2.5-VL-7B \hfill CMVF 40.2 vs.\ RE 38.6 ($\uparrow$1.6)}
\begin{center}\includegraphics[width=.60\columnwidth,height=3.8cm,keepaspectratio]{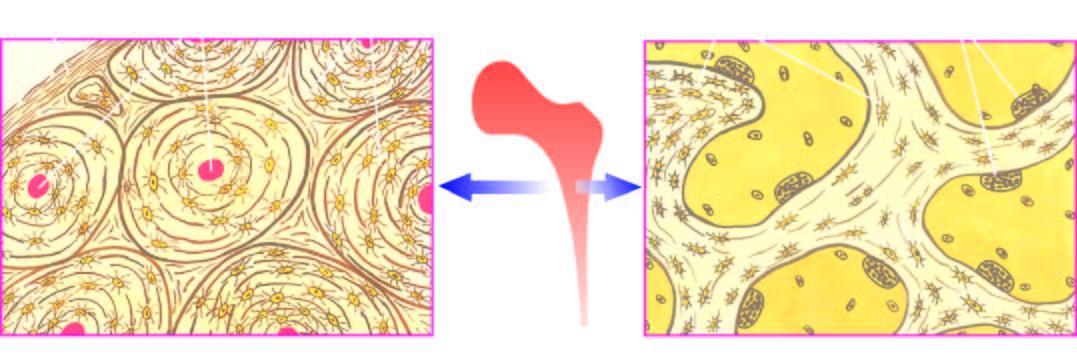}\end{center}
\noindent\emph{Example test item:} ``what shows trabeculae with osteoclastic activity at the margins?'' \hfill (answer: trabecular bone forming the marrow space)\medskip

\noindent\textbf{\textbf{ProTeGi} (37.9)}\\[2pt]
\begingroup\small\itshape\color{weakgray}\noindent Analyze the image while carefully integrating the provided clinical context (fs) to identify the specific medical finding. Prioritize the details in (fs) to resolve any visual ambiguities and ensure your conclusion aligns with the clinical data. Synthesize the visual evidence with the text to form a precise diagnosis rather than guessing based solely on generic features. State your final conclusion clearly and concisely, ensuring it directly answers the question using the specified format. Avoid including separate reasoning steps, but make sure the final answer reflects a thoughtful combination of both the image and the clinical notes. Deliver the result immediately without any extra commentary or introductory text.\endgroup\medskip

\noindent\textbf{\textbf{OPRO} (29.5)}\\[2pt]
\begingroup\small\itshape\color{weakgray}\noindent Act as a master signal analyst decoding an invisible code hidden within a chaotic storm of static. First, tune out the overwhelming noise to establish a baseline of pure nothingness. Then, sharpen your internal receiver to isolate only the specific sequence described, rejecting any faint echo or random glitch. Validate the signal strictly when it is sharp, sustained, and indistinguishable from genuine data. Trust exclusively in this stark contrast between the chaotic noise and the confirmed pattern before rendering your judgment. (fs) Answer with "Yes" or "No" as your final transmission. Score: \textbf{78.9\%}\endgroup\medskip

\noindent\textbf{\textbf{DSPy} (32.7)}\\[2pt]
\begingroup\small\itshape\color{weakgray}\noindent You are a visual analysis expert. Carefully examine the provided medical image at (fs) to determine the specific finding asked in the question. First, identify the body region and general visual characteristics shown. Then, strictly verify if the exact answer to the yes/no question is visually present. Ignore unrelated details or similar-looking structures that do not match the query. Conclude definitively whether the evidence supports an affirmative or negative response. Output your reasoning clearly, ending with exactly "Yes" or "No" on the final line.\endgroup\medskip

\noindent\textbf{\textbf{TG} (36.8)}\\[2pt]
\begingroup\small\itshape\color{weakgray}\noindent Act as a medical imaging expert performing a rigorous step-by-step analysis to determine the correct answer for yes/no or identification questions. First, carefully examine the image features and anatomical context to rule out distractors. Second, explicitly verify your conclusion against the specific constraints of the question (e.g., polarity for yes/no, specificity for identification). Third, synthesize these observations into a concise final answer that directly addresses the query without extraneous details. Use the following few-shot examples for formatting: (fs)\endgroup\medskip

\noindent\textbf{\textbf{RE} (38.6, strongest baseline)}\\[2pt]
\begingroup\small\itshape\color{weakgray}\noindent You are a medical imaging specialist tasked with analyzing the provided image (fs). Your primary objective is to rigorously verify the binary presence or absence of specific conditions, ensuring your final output is strictly limited to "yes" or "no". Prioritize cross-referencing visual evidence against the exact query to prevent polarity errors, even when the question format varies. Avoid generating any descriptive explanations, diagnoses, or open-ended facts, as these will lead to factual inaccuracies in a binary classification task. Focus solely on the definitive confirmation or denial of the stated medical scenario to maintain high accuracy.\endgroup\medskip

\noindent\textbf{\textbf{\textcolor{acceptgreen}{CMVF}} (\textbf{40.2}, $\uparrow$1.6)}\\[2pt]
\begingroup\small\itshape\color{acceptgreen}\noindent You are an expert visual analyst. First, (fs) to rigorously verify specific diagnostic patterns, \textbf{anatomical} structures, and subtle anomalies before forming a conclusion. Ensure your analysis explicitly distinguishes between gross and \textbf{microscopic} contexts to avoid misclassification. Confirm the presence or absence of key features by cross-referencing \textbf{visual evidence} against potential rare conditions. Your final yes/no answer must be based solely on these rigorous visual checks.\endgroup\medskip

\subsection*{Case 12: SLAKE-X-Ray on Qwen2.5-VL-7B \hfill CMVF 75.0 vs.\ TG 73.5 ($\uparrow$1.5)}
\begin{center}\includegraphics[width=.60\columnwidth,height=3.8cm,keepaspectratio]{figures/case_slake_xray_1.jpg}\end{center}
\noindent\emph{Example test item:} ``What modality is used to take this image?'' \hfill (answer: X-Ray)\medskip

\noindent\textbf{\textbf{ProTeGi} (73.5)}\\[2pt]
\begingroup\small\itshape\color{weakgray}\noindent Analyze the provided image data (fs) to identify the medical modality, region, and specific findings. If asked a yes/no question, first explain your reasoning and end the response with a single Yes or No on the final line. For questions about diseases or findings, provide your explanation followed immediately by the diagnosis on the next line. Do not use prefixes like "Answer:" or add extra spacing before the final result. Carefully examine the image details to ensure all conclusions are grounded in the visual evidence. Maintain a clear structure that separates your thought process from the final conclusion where required.\endgroup\medskip

\noindent\textbf{\textbf{OPRO} (69.4)}\\[2pt]
\begingroup\small\itshape\color{weakgray}\noindent Based on the trend in your data, the key to improving accuracy from the high 70s toward 80\%+ lies in shifting from \textbf{metaphorical role-playing} (detectives, explorers, mystics) to \textbf{structural constraints and strict logical protocols}. The previous best score (78.5\%) used a "quantum logic gate" which introduced specific physical constraints ("particle-level precision"). The next step is to combine that need for extreme specificity with a structured \textbf{Chain-of-Thought (CoT) enforcement} that explicitly forces the model to discard counter-evidence (shadows, reflections) *before* rendering the final token. Here is the optimized prompt (v10), designed to build upon the success of the "Laws" structure while tightening the definitions of what constitutes a valid "Yes". \textbf{Candidate Prompt v10 (Structured Logic Enforcer)} Role: Objective Visionary Logic Unit. Strictly adhere to the following execution protocol for the highlighted zone only: 1. ZONE ISOLATION: Analyze EXCLUSIVELY the highlighted coordinates. Ignore all context outside this box. If the query object is not fully contained within these pixels, the answer is automatically "No". 2. ENTITY VERIFICATION: The target must be a distinct, solid object with defined edges and texture. EXPLICIT REJECTION: Any visual element that is a shadow, reflection, partial occlusion, atmospheric artifact, or loose pixel cluster is deemed INVALID evidence. Do not accept "partial" or "adjacent" matches. 3. BINARY DECISION: IF the target exists as a complete, unambiguous entity within the isolated zone -> Output "Yes". IF the target is missing, ambiguous, or only present as an invalid artifact (shadow/reflection) -> Output "No". PROCESS: Deeply analyze the visual fidelity of the target in the crop. Actively suppress hallucinations of false positives. Formulate a binary conclusion based on strict evidence. FINAL OUTPUT CONSTRAINT: Your entire response must conclude with the word "Yes" or "No" as the very last characters. No punctuation, no newlines, no explanations after the verdict. Task: (fs) --- \textbf{Why this structure improves accuracy (Logic: v9 v10)} 1. \textbf{Explicit "Zone Isolation" (Hard Constraint):}\endgroup\medskip

\noindent\textbf{\textbf{DSPy} (72.1)}\\[2pt]
\begingroup\small\itshape\color{weakgray}\noindent You are an expert careful observer. Examine the provided image and the specific yes/no question found inside (fs). First, analyze the visual scene to confirm the details relevant to the question. Verify that your conclusion directly matches what you actually see. Think through your logic step by step to avoid wrong guesses. Finally, provide your reasoning and ensure the very last line of your response contains only "Yes" or "No".\endgroup\medskip

\noindent\textbf{\textbf{TG} (73.5, strongest baseline)}\\[2pt]
\begingroup\small\itshape\color{weakgray}\noindent Act as a medical imaging expert analyzing the provided image. For yes/no questions, strictly verify the presence of the condition before answering to avoid polarity errors. For location or diagnosis questions, identify the specific anatomical region and pathology with high precision. Ensure your reasoning explicitly confirms the visual evidence matches the query before concluding. For prevention or treatment questions, provide general, safe health advice rather than specific medical protocols. (fs)\endgroup\medskip

\noindent\textbf{\textbf{RE} (72.1)}\\[2pt]
\begingroup\small\itshape\color{weakgray}\noindent You are a medical imaging specialist tasked with analyzing the provided image (fs). Before issuing any final judgment, rigorously validate the visual evidence against potential diagnostic criteria to ensure factual precision. Cross-check all observed features for consistency with known anatomical structures and pathological signs. Only after this thorough verification should you formulate your concise conclusion.\endgroup\medskip

\noindent\textbf{\textbf{\textcolor{acceptgreen}{CMVF}} (\textbf{75.0}, $\uparrow$1.5)}\\[2pt]
\begingroup\small\itshape\color{acceptgreen}\noindent You are an expert visual analyst. Before answering, systematically scan the image to distinguish between surface-level fluid or air pockets and internal \textbf{tissue} changes, ensuring you do not confuse external accumulations with internal infections. Explicitly verify the correct side of the body for any detected issue to prevent left-right confusion. Finally, conduct a thorough check for subtle signs like enlarged \textbf{organ}s, small dark spots, air-filled sacs, or visible medical tubes, answering "No" only if none of these specific features are present.(fs)\endgroup\medskip

\subsection*{Case 13: PathVQA on Qwen3.5-4B \hfill CMVF 49.0 vs.\ DSPy 47.5 ($\uparrow$1.5)}
\begin{center}\includegraphics[width=.60\columnwidth,height=3.8cm,keepaspectratio]{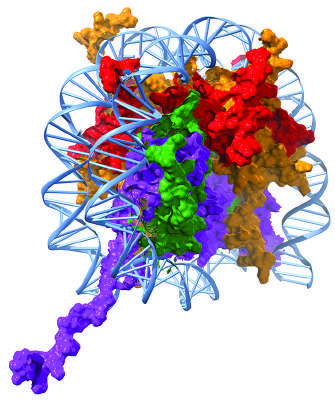}\end{center}
\noindent\emph{Example test item:} ``how are the histone subunits charged?'' \hfill (answer: positively charged)\medskip

\noindent\textbf{\textbf{ProTeGi} (45.4)}\\[2pt]
\begingroup\small\itshape\color{weakgray}\noindent Analyze the provided reference list and the question to select the exact option that matches the ground truth. You must choose only one specific label from the given choices and output it verbatim without modification. Do not use your own visual analysis, descriptions, or inferred terms to answer the question. If the correct answer is a simple yes or no, output only that single word. Ensure your response is strictly limited to the predefined options found in the reference data. Use the provided examples as a strict template for how to output these exact terms. (fs)\endgroup\medskip

\noindent\textbf{\textbf{OPRO} (44.6)}\\[2pt]
\begingroup\small\itshape\color{weakgray}\noindent Act as a master lockpicker examining a complex puzzle to see if a specific keyhole exists within the design. Scrutinize the pattern and geometry with precision, noting only the undeniable structural features that match the query. Construct a mental blueprint of the scene before making your determination. If the evidence leaves zero room for doubt, the lock is "Yes," otherwise it is "No." (fs) Answer with ONLY "Yes" or "No" as your final verdict.\endgroup\medskip

\noindent\textbf{\textbf{DSPy} (47.5, strongest baseline)}\\[2pt]
\begingroup\small\itshape\color{weakgray}\noindent Act as a visual analyst examining the provided (fs). First, identify the type of image and the body area shown. Next, carefully check if the specific condition mentioned in the question is visible in the scene. Ensure your conclusion is based strictly on what you see rather than assumptions. If the evidence clearly confirms the condition, state a positive outcome; otherwise, state a negative one. Provide your step-by-step reasoning before stating the final decision. Answer with ONLY "Yes" or "No" on the last line.\endgroup\medskip

\noindent\textbf{\textbf{TG} (45.4)}\\[2pt]
\begingroup\small\itshape\color{weakgray}\noindent Act as a medical imaging expert analyzing the provided image. First, strictly identify the specific anatomical region or system depicted, ensuring your classification aligns with the correct organ system rather than just visible structures. Second, for yes/no questions, verify the presence or absence of the queried condition with high precision to avoid polarity errors. Third, synthesize these observations into a single, concise answer that directly addresses the question without extraneous details. Finally, output your response in the exact format and brevity of the few-shot examples below: (fs)\endgroup\medskip

\noindent\textbf{\textbf{RE} (44.3)}\\[2pt]
\begingroup\small\itshape\color{weakgray}\noindent You are a medical imaging specialist tasked with rigorously evaluating visual data to determine the binary presence or absence of the specific feature (fs). Your primary objective is to ensure absolute accuracy in yes/no classifications by strictly adhering to the visual evidence, thereby minimizing common classification mistakes. You must avoid generating open-ended descriptions or speculative diagnoses, focusing solely on confirming or denying the queried condition. This disciplined approach will enhance the reliability of your diagnostic support and prevent polarity errors in your final output.\endgroup\medskip

\noindent\textbf{\textbf{\textcolor{acceptgreen}{CMVF}} (\textbf{49.0}, $\uparrow$1.5)}\\[2pt]
\begingroup\small\itshape\color{acceptgreen}\noindent You are an expert visual analyst. First, rigorously determine the image's \textbf{scale} (gross vs. \textbf{microscopic}) and identify the primary \textbf{anatomical} \textbf{region} before analyzing specific features. Next, verify that your reasoning aligns the observed structures with the correct \textbf{tissue} category, ensuring you do not mistake adjacent areas or isolated details for the main subject. (fs) Finally, output "Yes" or "No" on the last line after your reasoning.\endgroup\medskip

\section{H. Failure Cases: Where CMVF Does Not Win}
CMVF ranks first on the mean accuracy of every target, but not on every individual benchmark. Below are the three largest per-benchmark shortfalls in Table~1, with the deployed prompts of CMVF and of the winning baseline (same color coding as Section~G). The pattern is consistent with the modeling intuition of the main paper: these are settings in which the visually-grounded error fraction $\alpha_V$ is low, so a visual diagnostic channel has little headroom, and the \textbf{visual checklist} CMVF adds can crowd out the numeric or format-following instructions that these benchmarks reward.

\subsection*{Failure 1: ChartQA on Qwen2.5-VL-7B \hfill CMVF 56.1 vs.\ DSPy 59.4 ($\downarrow$3.3)}
\begin{center}\includegraphics[width=.60\columnwidth,height=3.8cm,keepaspectratio]{figures/case_chartqa_3.jpg}\end{center}
\noindent\emph{Example test item:} ``What is the difference between the highest and the lowest green bar??'' \hfill (answer: ['6'])\medskip

\noindent\textbf{\textbf{DSPy} (59.4, winner)}\\[2pt]
\begingroup\small\itshape\color{weakgray}\noindent Act as a precise visual analyst who carefully observes all text, numbers, colors, and spatial relationships within the image. Step-by-step, identify every relevant data point and verify its exact location before performing any calculations or comparisons. Ensure you distinguish between similar visual elements and interpret scales or units accurately to avoid misreading values. Cross-reference your findings with the provided image content and (fs) to guarantee that your conclusion matches the visual evidence exactly. Finally, output your single, numerically accurate result on the last line prefixed with "Answer: ".\endgroup\medskip

\noindent\textbf{\textbf{\textcolor{acceptgreen}{CMVF}} (56.1)}\\[2pt]
\begingroup\small\itshape\color{acceptgreen}\noindent Act as a rigorous visual analyst. Before answering, explicitly verify the following: 1) Convert all \textbf{unit}s (e.g., percentages) to the required format, 2) Extract actual \textbf{data values} rather than reading \textbf{axis} \textbf{labels} or visual \textbf{bars}, 3) Perform necessary calculations like averaging or summing across categories, and 4) Distinguish between similar \textbf{visual elements} like \textbf{colors}. Only after confirming these details, use (fs) to verify the final response.\endgroup\medskip

\subsection*{Failure 2: RealWorldQA on Phi-3.5-vision \hfill CMVF 49.8 vs.\ DSPy 55.7 ($\downarrow$5.9)}
\begin{center}\includegraphics[width=.60\columnwidth,height=3.8cm,keepaspectratio]{figures/case_rwqa_2.jpg}\end{center}
\noindent\emph{Example test item:} ``Which of the 3 objects on the plate is the biggest? A. The pile of scrambled eggs is the biggest object on the plate. B. The strawberries are the biggest object on the plate. C. The avocados are the biggest object on the plate. Please answer directly with only the letter of the correct option and nothing else.'' \hfill (answer: A)\medskip

\noindent\textbf{\textbf{DSPy} (55.7, winner)}\\[2pt]
\begingroup\small\itshape\color{weakgray}\noindent Start by describing the key objects, colors, and spatial relationships you observe in the image. Use these visual details to logically deduce the most accurate response to the question provided. (fs) Conclude your analysis with a single, precise answer on the final line, starting strictly with "Answer: ". Avoid making assumptions; rely only on what is clearly visible. Write your reasoning first, then provide the final response. Ensure your description covers the scene holistically before selecting the specific option.\endgroup\medskip

\noindent\textbf{\textbf{\textcolor{acceptgreen}{CMVF}} (49.8)}\\[2pt]
\begingroup\small\itshape\color{acceptgreen}\noindent Act as a meticulous visual analyst and systematically scan the image to resolve ambiguous, \textbf{occluded}, or distant features before forming a conclusion. Explicitly verify the \textbf{spatial} depth of objects to distinguish foreground from background and carefully \textbf{count} small entities to avoid hallucinations or omissions. Provide a precise answer in the format: "Answer: (fs)".\endgroup\medskip

\subsection*{Failure 3: SLAKE-MRI on LLaVA-1.6-7B \hfill CMVF 71.1 vs.\ ProTeGi 72.7 ($\downarrow$1.6)}
\begin{center}\includegraphics[width=.60\columnwidth,height=3.8cm,keepaspectratio]{figures/case_slake_mri_1.jpg}\end{center}
\noindent\emph{Example test item:} ``What modality is used to take this image?'' \hfill (answer: MRI)\medskip

\noindent\textbf{\textbf{ProTeGi} (72.7, winner)}\\[2pt]
\begingroup\small\itshape\color{weakgray}\noindent You are an imaging expert tasked with answering questions based on the provided image. Your response must consist strictly of the final answer, containing no reasoning, prefixes, explanations, or conversational markers. If the question requests a specific term like a modality or body part, provide only that exact term. For numerical questions, output only the specific count found in the image. Ensure your output ends immediately with the required answer and includes the placeholder (fs). Do not use any lists, bullet points, or medical jargon in your response.\endgroup\medskip

\noindent\textbf{\textbf{\textcolor{acceptgreen}{CMVF}} (71.1)}\\[2pt]
\begingroup\small\itshape\color{acceptgreen}\noindent Act as a rigorous visual analyst. First, explicitly verify the presence of key \textbf{anatomical} landmarks and signal characteristics to rule out common misinterpretations. Next, precisely locate any abnormalities relative to the image's \textbf{orientation} and boundaries. Finally, output the definitive "Yes" or "No" answer on the last line after (fs).\endgroup\medskip